\documentclass[runningheads]{llncs}

\usepackage[mobile]{eccv}

\usepackage{eccvabbrv}

\usepackage{graphicx}
\usepackage{booktabs}
\usepackage{amsmath} 
\usepackage{algorithm} 
\usepackage{algpseudocode} 

\usepackage[accsupp]{axessibility}  

\algdef{SE}[SUBALG]{Indent}{EndIndent}{}{\algorithmicend\ }%
\algtext*{Indent}
\algtext*{EndIndent}

\usepackage[pagebackref,breaklinks,colorlinks]{hyperref}

\begin{document}

\title{\textcolor{orange}{MoMA}: \textcolor{orange}{M}ultim\textcolor{orange}{o}dal LL\textcolor{orange}{M} \textcolor{orange}{A}dapter for Fast Personalized Image Generation} 

\titlerunning{MoMA}

\author{Kunpeng Song\inst{1,2} \and Yizhe zhu\inst{1} \and Bingchen Liu\inst{1} \and Qing Yan\inst{1} \and Ahmed Elgammal\inst{2} \and Xiao Yang\inst{1}}

\authorrunning{ }

\institute{ByteDance \and Rutgers University}


\maketitle

\begin{abstract}
In this paper, we present MoMA: an open-vocabulary, training-free personalized image model that boasts flexible zero-shot capabilities. As foundational text-to-image models rapidly evolve, the demand for robust image-to-image translation grows. Addressing this need, MoMA specializes in subject-driven personalized image generation. Utilizing an open-source, Multimodal Large Language Model (MLLM), we train MoMA to serve a dual role as both a feature extractor and a generator. This approach effectively synergizes reference image and text prompt information to produce valuable image features, facilitating an image diffusion model. To better leverage the generated features, we further introduce a novel self-attention shortcut method that efficiently transfers image features to an image diffusion model, improving the resemblance of the target object in generated images. Remarkably, as a tuning-free plug-and-play module, our model requires only a single reference image and outperforms existing methods in generating images with high detail fidelity, enhanced identity-preservation and prompt faithfulness. We commit to making our work open-source, thereby providing universal access to these advancements. 
\href{https://moma-adapter.github.io/}{Project page}
  
  \keywords{image generation \and multimodal \and personalization \and LLM}
\end{abstract}

\section{Introduction}

\begin{figure}
\centering
\includegraphics[width=\linewidth]{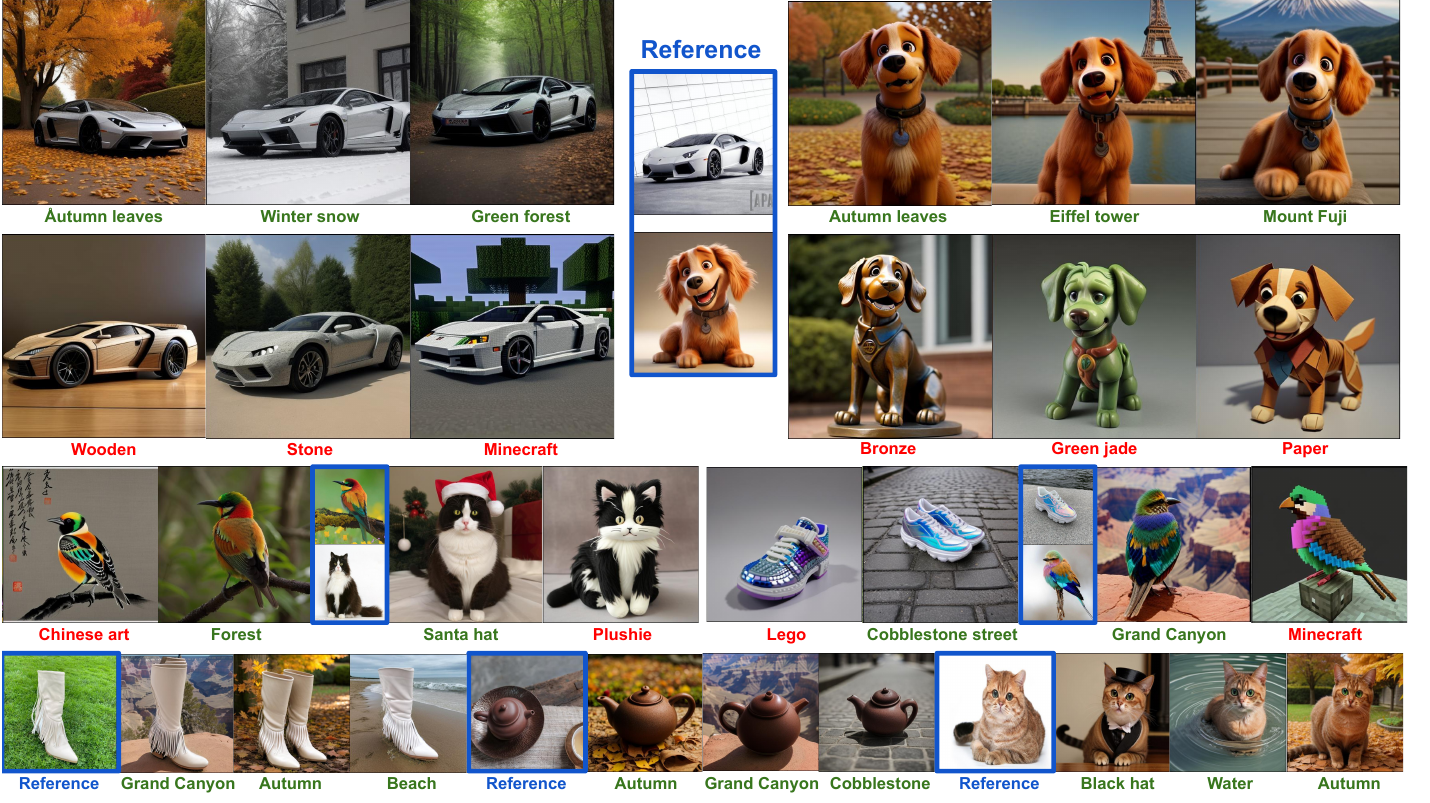} 
\caption{Example images generated by our open-vocabulary personalization model.  without tuning. With just one image of a subject (circled in blue), our model can generate text-aligned, identity-preserved new images of the same subject with only a single forward pass. Our model supports both re-contextualization where the same subject is located in a new environment, as shown in green, and changing the texture of the subject itself, as shown in red.}
\label{fig:hero}
\end{figure}

\label{sec:intro}
Image generation technology has seen remarkable advancements with the emergence of large-scale text-to-image diffusion models such as GLIDE\cite{nichol2021glide}, DALL-E 2\cite{ramesh2022hierarchical}, Imagen\cite{saharia2022photorealistic}, Stable Diffusion \cite{rombach2022high}, eDiff-I\cite{balaji2022ediffi}. These models enable users to generate vivid images from a diverse set of text prompts. Despite their effectiveness, textual descriptions often fall short in expressing detailed visual features, leading to the rise of image-conditioned generation works like Kandinsky\cite{razzhigaev2023kandinsky}, Stable Unclip\cite{rombach2022high,stableunclip}, which utilize images as inputs to create variations that maintain the visual components of the reference.

A natural progression in this field is subject-driven generation or image personalization. Initial efforts in this domain involve inverting input images into textual representations and employing learnable text tokens to denote target concepts. For instance, DreamBooth\cite{ruiz2022dreambooth} fine-tunes the entire diffusion model to learn a unique identifier for specific subjects. Textual-Inversion\cite{gal2022textual} inverts the input images to a unique textual embedding and learns the embedding-image mapping during finetuning. Subsequent approaches, such as Custom Diffusion\cite{kumari2022customdiffusion}, have optimized this process by focusing on tuning only cross-attention layers to reduce computational costs. Techniques like Low-Rank Adaptation (LoRA)\cite{hu2021lora} and SVDiff\cite{han2023svdiff} further minimized trainable parameters, enhancing fine-tuning efficiency. However, these methods, regardless of their accuracy, require extensive resources for per-instance tuning and model storage, limiting their practical application.

Tuning-free approaches have gained prominence for addressing these limitations. For example, IP-Adapter\cite{ye2023ip-adapter} leverages a unique cross-attention mechanism to differentiate text and image features, facilitating the injection of reference images as visual prompts. Nevertheless, it faces notable constraints, particularly in modifying textures. Methods like InstantID\cite{wang2024instantid}, InstantBooth\cite{shi2023instantbooth}, and PhotoVerse\cite{chen2023photoverse}, while maintaining identity coherence, are confined to specific domains such as human faces or cats. Recent innovations have employed transformers to integrate visual concepts with text prompts, as seen in BLIP-Diffusion\cite{li2024blip} and KOSMOS-G\cite{pan2023kosmos}, which extract identity information and combine it with text prompts via cross-attention. These approaches, although effective in texture modification, often result in significant detail errors in a tuning-free setting and require extra tuning for optimal outcomes with target objects.

In response to these challenges, this paper introduces a novel, open-vocabulary, and tuning-free image personalization model that excels in detail fidelity, object identity resemblance, and coherent textual prompt integration. Our model harnesses the capabilities of Multimodal Large Language Models (MLLMs) to seamlessly blend text prompts with visual features of the target object, enabling alterations in both the background context and object texture. In addition, our newly proposed self-attention shortcut significantly enhances the detail quality with minimal computational overhead.

Built upon Stable Diffusion\cite{rombach2022high} and LLaVA\cite{liu2023llava,liu2023improvedllava,liu2024llavanext}, MoMA has been rigorously evaluated on various tasks with a wide array of images and dynamic prompts:
\begin{itemize}

\item For exact-object recontextualization tasks, it demonstrates superior detail accuracy and faithfulness to the target object across varied backgrounds. 
\item For texture modification tasks, our method adeptly alters the texture of target objects as dictated by text prompts while preserving unmentioned visual features. 
\item Our model achieves the above performance through extensive pre-training, eliminating the need for tuning at the evaluation stage.
\end{itemize}

\section{Related work}
\subsection{Text-to-Image Diffusion Models.}
Text-to-image diffusion models have become the forefront of image generation technology, garnering widespread interest for their exceptional capabilities in recent years. These models typically utilize a pre-trained language decoder to transform text prompts into latent representations that guide the diffusion process for generating or editing images. Notable models like GLIDE\cite{nichol2021glide} and DisCo\cite{wang2023disco} leverage text-guided diffusion architectures and CLIP\cite{radford2021learning} guidance to enhance the fidelity and relevance of the generated images. Similarly, Stable Diffusion\cite{rombach2022high} stands out by executing the diffusion process in latent image space, thus significantly lowering computational demands. It was further advanced by Stable Diffusion XL(SDXL)\cite{podell2023sdxl}, which introduced a larger UNet and an additional text encoder for improved textual influence on the images. Also, diffusion models have shown remarkable efficiency in capturing data distributions for image synthesis, with applications expanding by integrating transformer-based architectures\cite{peebles2023scalable}. The advent of text-guided image synthesis, mainly through models like Stable Diffusion, highlights significant advancements in achieving top-tier results in text-to-image synthesis tasks. Stable Diffusion, a prominent latent diffusion model, operates within a latent space defined by a pre-trained autoencoder, enabling efficient handling of semantic features and visual patterns for enhanced image synthesis.

\subsection{Personalized Image Synthesis.}
Recently, personalization has become an emerging factor in the vision and graphics community. Previous researchers have explored optimization-based approaches, such as Textual Inversion\cite{gal2022textual} and DreamBooth\cite{ruiz2022dreambooth}. Later works found it's sufficient to just update cross-attention modules in the diffusion unet. Various parameter efficient optimization methods were then developed to further speed up tuning, such as LoRA\cite{hu2021lora} and SVDiff\cite{han2023svdiff}.

The drawback for these methods is they require fine-tuning for each new reference image, which is computationally expensive and time-consuming. More recent efforts attempt to get rid of per-object tuning such as \cite{wei2023elite, jia2023taming,shi2023instantbooth}, which pre-train the diffusion model on domain-specific images, such as cat and human face images. These models provide class-specific prior for generation thus require less tuning within that domain. However, they are constrained to the trained class and are not able to generalize to other subjects. Another category of works \cite{chen2023anydoor, avrahami2023break,li2024blip} focuses on more general open-vocabulary data. AnyDoor\cite{chen2023anydoor} and BreakTheScene\cite{avrahami2023break} generate new images of the same object under various backgrounds but fail in changing textures. More related to our work, BLIP-Diffusion\cite{li2024blip} uses a pre-trained transformer feature extractor and works on a wide range of subjects, however, its results contain abundant details mistakes and require few-step tuning to achieve good quality results. We posit that this is due to the lengthy information path negatively impacting the quality of image
features.  IP-adapter\cite{ye2023ip-adapter}, as a lightweight plug-and-play model, directly injects the visual feature of the reference image into the UNet and achieves promising performance. However, the migrated image features can hardly interact with target prompts. This traps into a trade-off between prompt fidelity and image fidelity, especially when the prompt requests for drastic context changing or texture editing: higher strength results in better subject details at the cost of worse prompt faithfulness.



Nevertheless, these methods all suffer from achieving identity preservation, edibility, generalization ability and high fidelity simultaneously. Our method, however, is able to make progress in those key directions within the single input image and tuning-free domain.

\section{Method}

In this section, we first introduce some preliminaries about text-to-image diffusion models and multmodal LLMs. Then, we depict in detail the motivation and the design of the proposed multimodal LLM adapter.

\begin{figure}
\centering
\includegraphics[width=\linewidth]{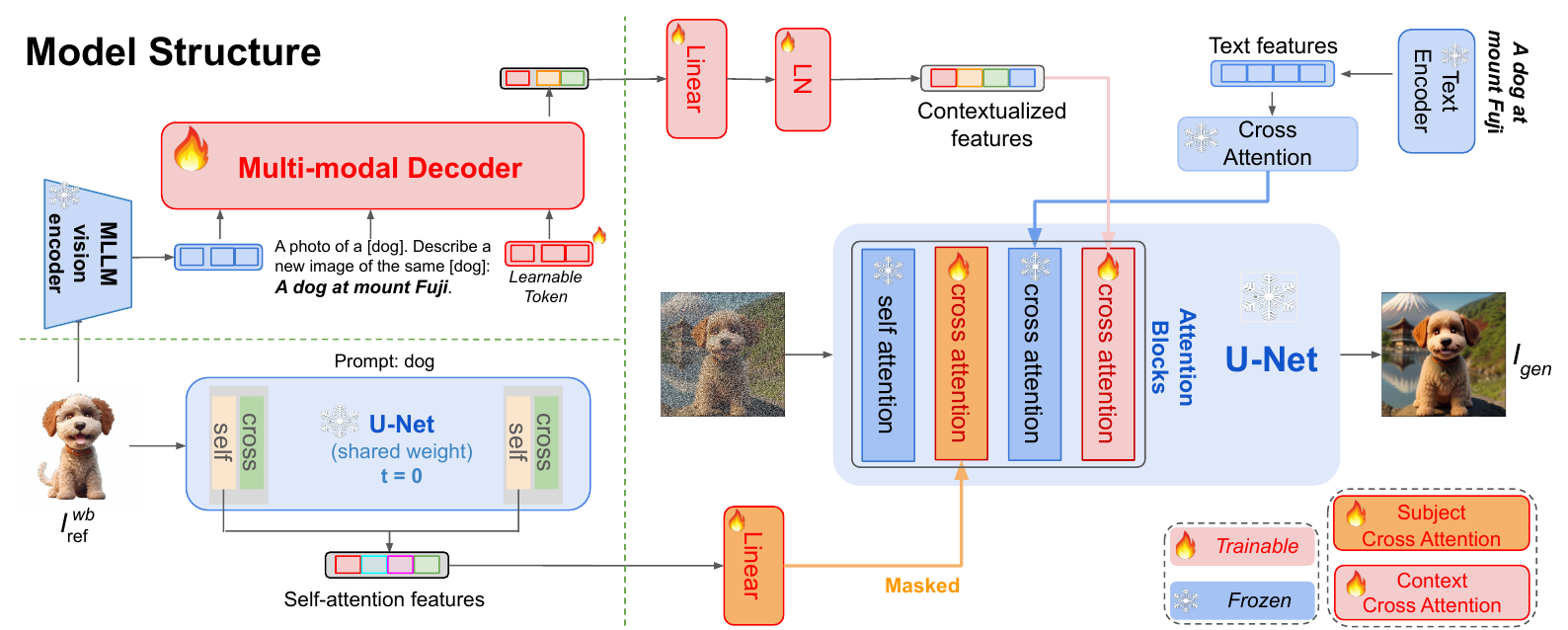} 
\caption{Model structure. (1) On top-left, we adopt a generative multimodal image decoder to extract semantic features and modify them by the target prompt. These features are projected to text space and then injected into a pretrained frozen UNet with decoupled context cross-attentions as illustrated in light red. (2) On bottom-left, to further improve detail accuracy, we forward the clear reference image ($t=0$) to the same UNet and extract self-attention features. These fine-grained features contain detailed information about the subject and are injected into UNet through decoupled object cross-attention layers as illustrated in orange. (3) The model is trained using a two-staged training pipeline: we first train the multimodal decoder (multimodal generative learning), then jointly optimize newly added attention modules in UNet.}
\label{fig:model_structure}
\end{figure}

\subsection{Preliminaries.}
\subsubsection{Text-to-Image Diffusion Models}
Text-to-image diffusion models generate images that align with the textual description by gradually denoising a random sample drawn from a Gaussian distribution. Our work is established on text-to-image Latent Diffusion Models (LDM), which perform the diffusion process in latent space, making it more practical and computationally efficient.  Given an input image $x$, LDM first extracts the latent feature $z  =E(x)$ with a well-trained encoder. During training, the noisy latent variables $z_t$ is obtained
by gradually adding noises to $z$ for $t$ steps, LDM optimizes the following objective:

\begin{equation}\label{eq:1}
\mathcal{L} = \mathbb{E}_{z_t, t, C, \varepsilon \sim \mathcal{N}(0,1)}\left[\left\| \varepsilon - \varepsilon_\theta(z_t, t, C) \right\|_2^2\right], 
\end{equation}


\noindent where $C$ denotes the textual embedding of prompts extracted by a pre-trained CLIP text encoder.

LDM is commonly parameterized as an UNet model. We employ the pre-trained Stable Diffusion as the LDM, where the UNet model has cross-attention and self-attention layers in different resolutions. Given the image features \( Z \) as query and the textual embedding \( C \) as key, the output of cross-attention \( Z' \) can be defined by the following equation:

\begin{equation}
Z' = Attn(Q, K, V), 
\end{equation}


\noindent where \( Q = ZW_q \), \( K = CW_k \), \( V = CW_v \) are the query, key, and values matrices of the attention operation respectively, and \( W_q \), \( W_k \), \( W_v \) are the weight matrices of the trainable linear projection layers.

\subsubsection{Multimodal LLM (MLLM)}
The field of natural language processing (NLP) has undergone a revolutionary shift with the emergence of large language models (LLMs). These models, distinguished by their comprehensive training across varied datasets, demonstrate extraordinary proficiency in a range of linguistic tasks. Benefiting from robust pre-training methodologies, pioneering models, such as ChatGPT-Vision\cite{achiam2023gpt}, Mini-GPT4\cite{zhu2023minigpt}, CogVLM\cite{wang2023cogvlm}, and LLaVA\cite{liu2023llava}, have set the stage for more sophisticated iterations in the multimodal field. These iterations notably integrate visual tasks into the LLM framework, and have proven to be exceptional in vision-related tasks, including vision-question-answering, visual-grounding, and visual-segmentation.

Among these MLLMs, LLaVA \cite{liu2023llava} stands out as an open-source Large Language and Vision Assistant, synergizing a vision encoder with an LLM for comprehensive visual and language understanding. LLaVA capitalizes on the strengths of both the pre-trained LLM and a vision transformer image encoder. This model skillfully processes images in tandem with language instructions, delivering responses in natural language, thereby bridging the gap between vision and linguistic comprehension.

\subsection{Methodology}


We present MoMA, a multimodal LLM adapter enhanced by fine-grained feature transfer. The overall architecture is demonstrated in Figure \ref{fig:model_structure}. Our method consists of three parts: (1) a generative multimodal decoder is utilized to extract image features from the reference image and edit it following the target prompt, yield the contextualized image feature; (2) in the meantime, we replace the background of the original image by white color, leaving only object pixels, leveraging the original UNet's self-attention layers to extract the object image feature; (3) finally, during the new image generation process, we injected the contextualized image features and the object image features into the UNet diffusion model with the dedicatedly trained context-cross-attention layers and object-cross-attention layers, respectively.

\subsubsection{Multimodal Generative Image-feature Decoder}


We introduce a multimodal generative image-feature decoder, which actively generates target image features by combining visual information from the reference image and textual information from text prompt. Practically, we adapt a pre-trained MLLM, specifically LLaVA-7B, to serve as our generative multimodal decoder.
As shown in \cref{fig:model_structure} upper-left branch, given a reference image $I_{ref}$ and its object mask $M_{ref}$, we get a white-background reference image by $I^{wb}_{ref} = I_{ref} * M_{ref}$.
We construct an instruction sequence as the input to MLLM: " $<f_{ref}>$ An image of $<label>$. Describe $<P_{\text{tgt}}>$", where $label$ is the subject keyword (e.g. $cat$, $car$, etc.), and $P_{tgt}$ the target prompt. 
A learnable token is appended at the end of the instruction sequence. After forwarding the MLLM, the embedding corresponding to this learnable token is the output of our multimodal image-feature decoder. We call it decoder as, intuitively, it is trained to combine visual features with the target prompt in a generative manner and output an image embedding. By design, the MLLM image-feature decoder edits the background-excluded image feature of $I^{wb}_{ref}$ following a background-included target prompt $P_{tgt}$ that describes an entire image. 

The generated image feature from the multimodal image-feature decoder is then converted into a sequence of embedding in $\mathbb{R}^{768}$ with length N (we use N = 4 in this work) through a linear layer. 
Inspired by IP-Adapter\cite{ye2023ip-adapter}, the embedding sequence is then integrated into the pre-trained UNet model with decoupled cross-attention as shown in \cref{fig:model_structure} upper-right branch. 

\subsubsection{Self-Attention Feature Transfer}

\begin{figure}[t]
\centering
\includegraphics[width=\linewidth]{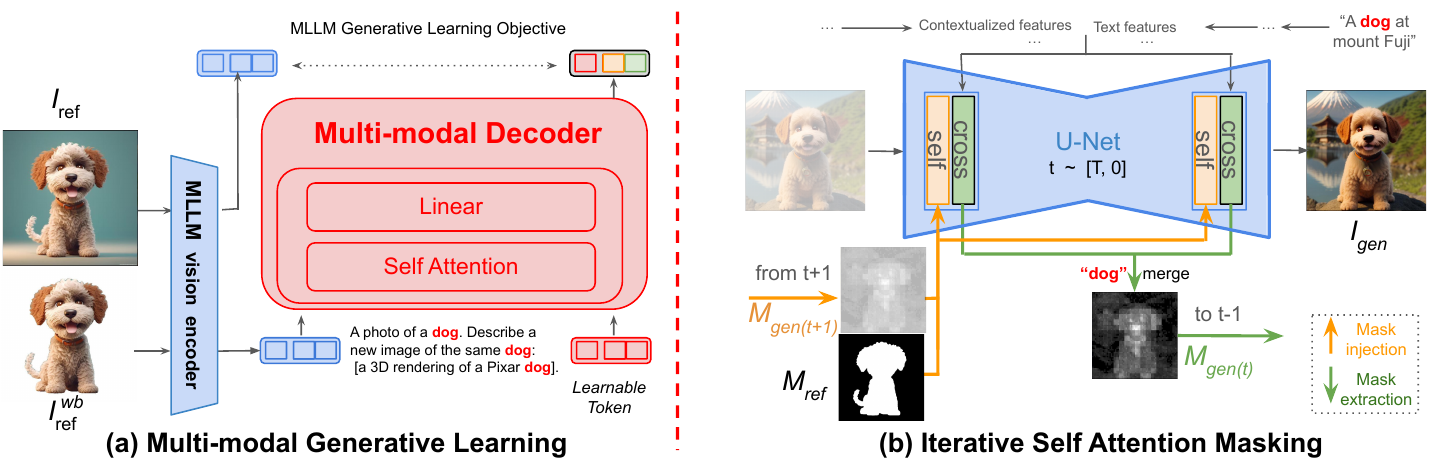} 
\caption{Multimodal Generative Learning and iterative Self-Attention Masking}
\label{fig:model_structure_2}
\end{figure}

To further enhance the detail faithfulness, we involve image self-attention features and apply a masking mechanism. Specifically, the same pre-trained UNet is leveraged as the self-attention feature extractor. As shown in \cref{fig:model_structure} lower branch, $I^{wb}_{ref}$ is forwarded through the diffusion UNet with $t=0$ as timestep and $label$ as text condition. Features at each self-attention layer are collected and transferred into the main UNet model by the adapted modules with decoupled self-attention.
        
The self-attention feature transfer is an effective information shortcut as the extracted feature $c_{i}$ carries fine-grained details. However, directly applying it will cause interference between the backgrounds of $I^{wb}_{ref}$ and $I_{gen}$. To address this issue, we present a self-attention masking procedure. Ideally, 
we want only the features of the foreground in $I^{wb}_{ref}$ to be injected into the foreground of $I_{gen}$. The features of the background in $I^{wb}_{ref}$ should be eliminated and the background of $I_{gen}$ should remain unaffected by the self-attention feature transfer. We apply a masking mechanism using the reference image mask $M_{ref}$ and the generated image mask $M_{gen}$. The output of our modified self-attention is:


\begin{equation}
Z_{new}=Attn(Q,K,V) + \lambda \cdot Attn(Q,K',V',M_{ref}) \cdot M_{gen} \cdot \beta
\end{equation}

\noindent where $\lambda$ is a learnable parameter. \( K'\) and \( V'\) are the key and values calculated from the extracted self-attention feature $c_{i}$ by \( K' = c_{i}W'_k \) and \( V' = c_{i}W'_v \). Here, \( W'_k \), \( W'_v \) are the weight matrices of the newly introduced decoupled subject-cross-attention projections. The reference image mask $M_{ref}$ is applied inside of $Attn$ in the form of the attention mask, and the generation mask $M_{gen}$ is applied through an element-wise product. $\beta$ is a strength scalar for additional controls.



During training, the model is optimized to reconstruct the background-included reference image $I_{ref}$. The white-background reference image $I^{wb}_{ref}$ and the target image $I_{ref}$ share the same mask, so $M_{ref}=M_{gen}$. During inference, the ground truth $M_{ref}$ is available but the ground truth $M_{gen}$ isn't. We use the cross-attention map corresponding to the subject $label$ to approximate $M_{gen}$. Specifically, as shown in \cref{fig:model_structure_2} (b), during each denoising step, the attention map of $label$ from each cross-attention layer is extracted and averaged into $M_{{gen(t)}}$. We use it to approximate $M_{gen}$ in the next denoising step.

\subsubsection{Multimodal Generative Learning and Diffusion Learning.}

Unlike previous works that extract the image features of the subject as it is, we generate image features that are well-modified following the target text prompt. Previous works, like IP-Adapter, inject image features into the cross-attention layers of the UNet without interacting with the target prompt. This is problematic, especially when the target prompt involves texture-changing the subject. On the other hand, our multimodal image-feature decoder imagines the full image given a white-background object image and a text prompt describing the full image. 
which dramatically improves model performance, especially in changing subject textures. It ensures the output preserves the identity of the target object while respecting the text prompt. To achieve the best model performance, we propose a two-staged pre-training strategy. 

First, we propose a \textbf{Multimodal Generative Learning Stage}, where we pre-train the multimodal image-feature decoder such that it learns to compose image features of the subject with the target prompt and output the CLIP embedding of the target image. To this end, we need to take advantage of the generative capability of the MLLM: while initially trained to generate text, we adapt it to generate image embeddings. As shown in Figure \ref{fig:model_structure_2} (a), $I^{wb}_{ref}$ is encoded by the MLLM vision encoder and combined with its caption $P_{ref}$, together with a learnable token, into a prompt instruction. This sequence is fed into the MLLM: a 15-layer transformer. The output of the learnable token is trained to match the CLIP image embedding of the original reference image $I_{ref}$. Once being well-trained, our MLLM will generate prompt-contexualized image embeddings. The loss function of this stage is formulated as:

\begin{equation}
\label{equation_loss}
\mathcal{L}_{\text{MLLM}} =  \left\|\text{MLLM}\left( \text{CLIP}\left( I^{wb}_{ref} \right), \text{P}_{ref}, \text{Token} \right) - \text{CLIP}\left( I_{ref} \right)\right\|_2^2
\end{equation}

Second, we design a \textbf{Diffusion Learning Stage} that faithfully converts the contextualized image embeddings to an image. During this stage, we freeze MLLM and pre-trained diffusion model and optimize only the decoupled subject and contextual attentions and their linear mappings. The model is trained on the OpenImage dataset, using the same training objective as shown in \cref{eq:1}.

Classifier-free guidance (CFG)\cite{ho2022classifier,ye2023ip-adapter} improves diffusion generaton quality. However, we find it better to only enable it for the context-cross-attention side and not on the subject-cross-attention side. Specifically, in the second training stage, to enable CFG on the context-cross-attention side, we randomly 
replacing the contextualized feature with an all-zero image embedding.



\begin{figure}[t]
\centering
\includegraphics[width=\linewidth]{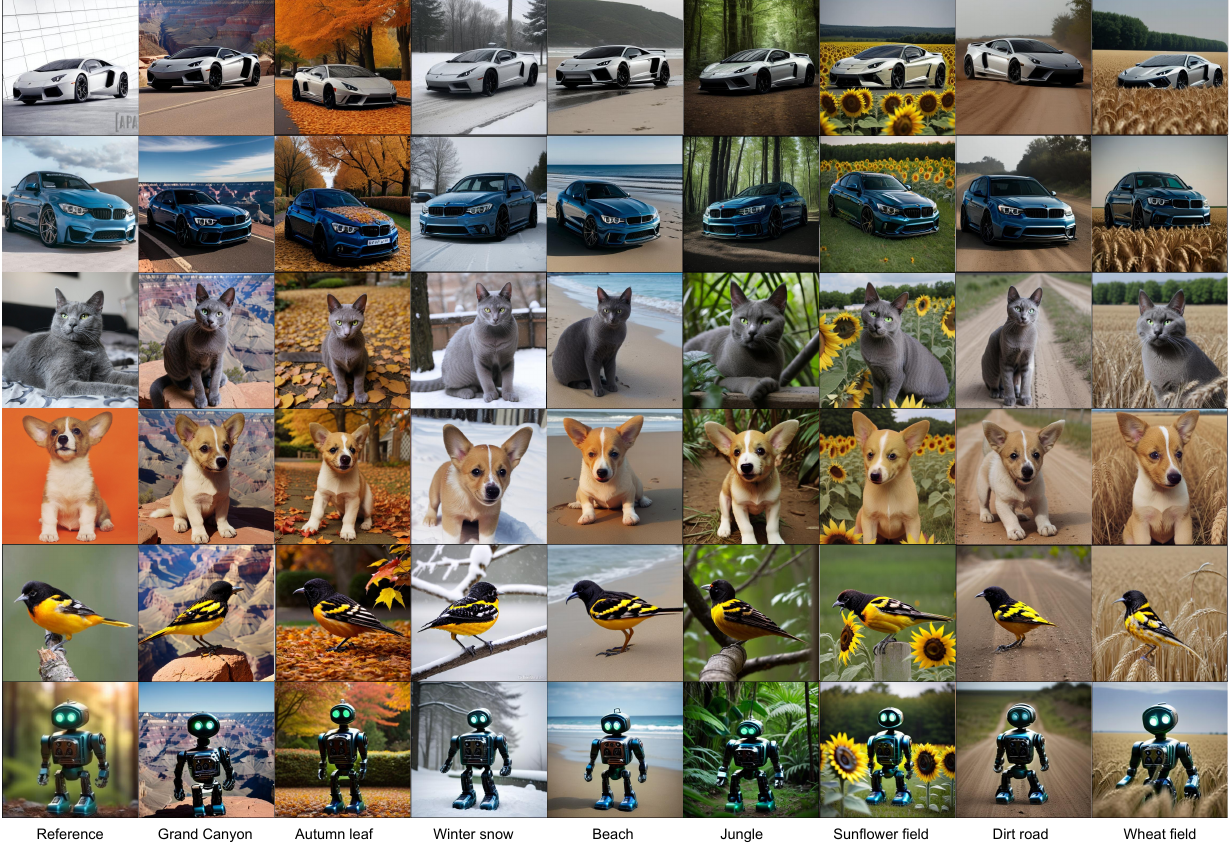} 
\caption{Zero-shot new context. Visualization of our generated samples for various images and prompts. Exact subject with different context.}
\label{fig:Recontextualization}
\end{figure}

\section{Experiments}
\subsection{Training and Implementation Detail} 
To train our model, we construct a dataset of 282K image/caption/image-mask triplets from the OpenImage-V7 \cite{kuznetsova2020open} dataset. We use BLIP-2 OPT6.7B to generate captions \cite{li2024blip} for the images, then remove human-related subjects and filter out color, shape, and texture keywords. We use the subject mask provided in OpenImage as $M_{ref}$. Evaluation images do not come with a mask, so we use SAM\cite{kirillov2023segany} to extract main objects and build masks thereafter. We use Stable Diffusion v1.5\cite{rombach2022high} with RealisticVision\cite{saharia2022photorealistic} checkpoint as our foundation diffusion model. We load LLaVA-7B as our MLLM decoder in stage-one training. In stage-two training, we load IP-Adapter\cite{ye2023ip-adapter} checkpoints to initialize our context cross-attention layers, and zero-initialize our object cross-attention layers. We evaluate the model using various images and prompts. 

\begin{figure}[t]
\centering
\includegraphics[width=1.0\linewidth]{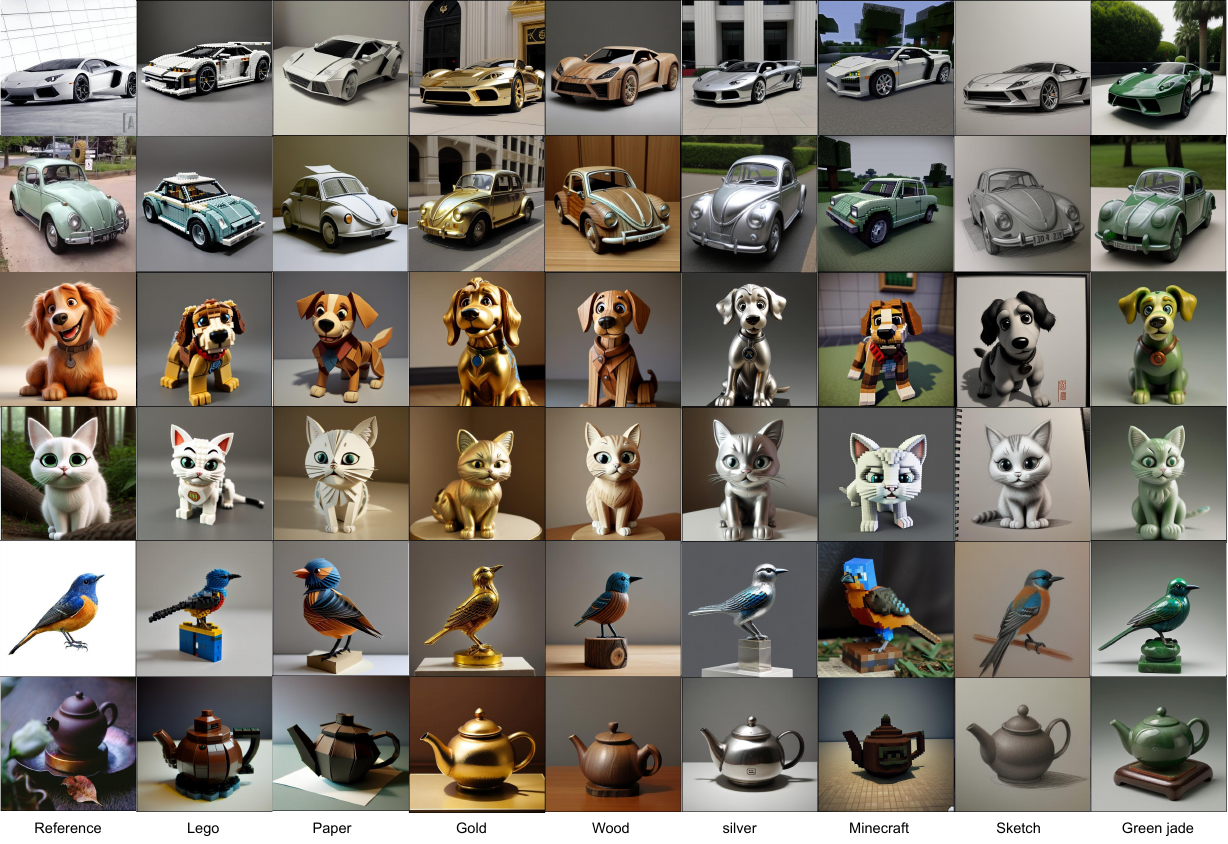} 
\caption{Zero-shot new texture. Visualization of our results with new texture, material, color and style. Our model correctly balances between prompt and image fidelity.}
\label{fig:texture}
\end{figure}

We present qualitative examples to illustrate the effectiveness of our model. In \cref{fig:Recontextualization}, the target prompts specify a novel contextual environment. Our model seamlessly generates a high-quality background while precisely situating the same object within this new setting. In \cref{fig:texture}, the prompts indicate a change in texture. Our model showcases its ability to render realistic textures in response to the textual cues, adeptly altering specified visual elements while leaving other identity aspects of the image unaffected.

\begin{figure}[t]
\centering
\includegraphics[width=\linewidth]{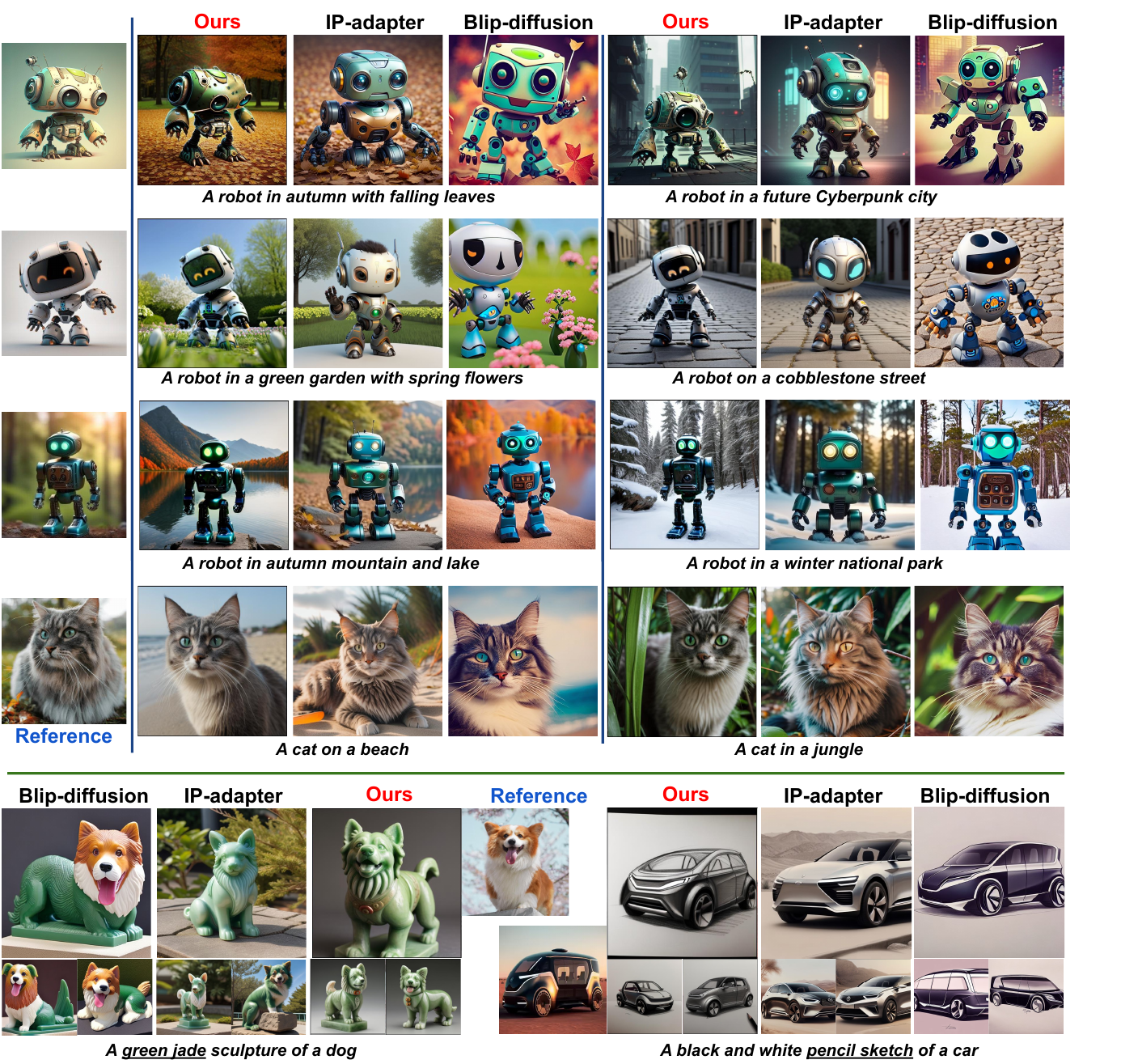} 
\caption{Zero-shot Qualitative Comparison. We share recontextualization in upper panel and texture editing in lower panel. Our results have significantly more accurate details for context editing and better balancing between prompt and image fidelity in texture editing.}
\label{fig:compare}
\end{figure}

\subsubsection{Qualitative Comparison.} We conducted a comparative analysis to evaluate the performance of our method against existing tuning-free open-vocabulary personalization approaches. For recontextualization, we use a variety of prompts and images to qualitatively assess our model. To ensure a fair comparison, we generated 50 random samples for each model and presented the highest-quality examples. The results, as illustrated in \cref{fig:compare} upper panel, show our method's proficiency in detail and background quality. Particularly, the green-marble eyes and facial details in cat images is notably refined. The backgrounds generated by our model also exhibit enhanced appeal, diversity, and realism, a benefit attributed to our masked cross-attention mechanism. In texture editing, analyzed in \cref{fig:compare} lower panel, our method consistently maintained the shape and contour of subjects while adapting them to eight different textures. This contrasts with baseline methods, which often struggled to balance the prompts and images effectively. Our results indicate a marked improvement in texture adaptation while preserving the integrity of the original subjects. We show more qualitative and uncurated results in Appendix B.




\subsubsection{Quantitative Comparison}
We use the Dreambooth\cite{ruiz2022dreambooth} dataset to conduct the quantitative evaluation. For a fair comparison, we generate 4 images conditioned on the image prompt for each dataset sample, resulting in 14k generated images for each method. In \cref{tab:my_label_qun}, we evaluate subject fidelity using DINO\cite{caron2021emerging} and CLIP-I\cite{radford2021learning} scores, and prompt-following ability using CLIP-T \cite{radford2021learning} scores. We disaggregate the CLIP-T score into two distinct scenarios: the recontextualization task denoted by CLIP-$I_{(c)}$, and the texture editing task denoted by CLIP-$I_{(t)}$. To evaluate CLIP-$I_{(t)}$, we introduce 10 new prompts specifically tailored to texture editing. We exclude CLIP-I and DINO scores from the texture editing experiments, as they could not serve as effective metrics in this scenario: a high DINO/CLIP-I score frequently doesn't equate to high quality; rather, it indicates a failure to alter the texture as intended. 

In \cref{tab:user}, we conduct user studies on recontextualization and texture editing. Our method shows a significant performance boost across these metrics, especially in prompt-following. We show a detailed breakdown of quantitative results and more information about human evaluation in Appendix B and C.

\begin{table}[htbp]
\centering
\begin{minipage}{0.53\textwidth} 
    \caption{Quantitative comparisons.}
    \label{tab:my_label_qun}
    \resizebox{\textwidth}{!}{
        \begin{tabular}{l|ccc|c}
        \hline
        Methods & DINO $\uparrow$ & CLIP-$T_{(c)}$ $\uparrow$ & CLIP-$I$ $\uparrow$  & CLIP-$T_{(t)}$\\
        \hline
        Real Images & 0.774 & 0.885 & -- & -- \\
        Textual Inversion & 0.569 & 0.255 & 0.780  & --\\
        Re-Imagen & 0.600 & 0.270 & 0.740  & --\\
        IP-Adapter & 0.612 & 0.330 & 0.793  & 0.319\\
        BLIP-Diffusion & 0.594 & 0.300 & 0.779  & 0.271\\
        Ours & \textbf{0.618} & \textbf{0.348} & \textbf{0.803}  & \textbf{0.335}\\
        \hline
        \end{tabular}
    }
\end{minipage}\hfill 
\begin{minipage}{0.47\textwidth} 
    \caption{Human Evaluations.}
    \label{tab:user}
    \resizebox{\textwidth}{!}{
        \begin{tabular}{lccc}
            \toprule
            Metric & Ours & IP-Adapter & BLIP-Diffusion \\
            \midrule
            Subject Fidelity $\uparrow$ & \textbf{0.550} & 0.416 & 0.034 \\
            Prompt Fidelity $\uparrow$ & \textbf{0.633} & 0.351 & 0.016 \\
            \midrule
            Texture Editing $\uparrow$ & \textbf{0.523} & 0.316 & 0.161 \\
            \midrule
            Overall Quality $\uparrow$ & \textbf{0.596} & 0.368 & 0.0366 \\
            \bottomrule
        \end{tabular}
    }
\end{minipage}
\end{table}

\begin{figure}
  \centering
  \includegraphics[width=1.0\textwidth]{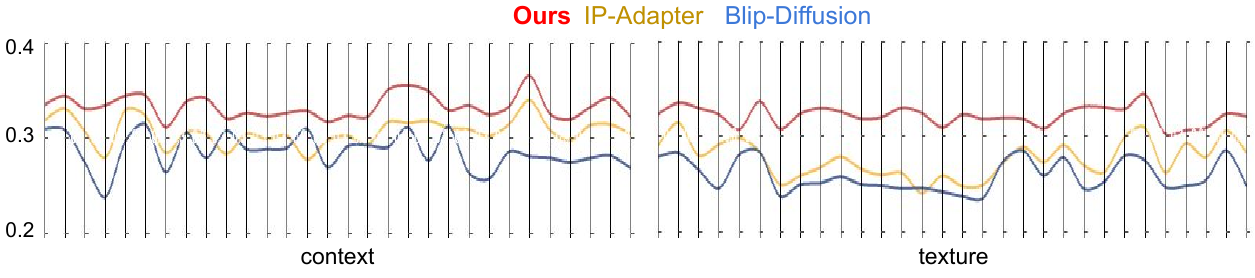} 
  \caption{Visualized CLIP-T for 30 subjects. The left chart is for the context editing task and the right is for texture editing. In each chart, one vertical axis represents one subject. Results from different models are colored differently. }
  \label{fig:visual}
\end{figure}

In \cref{fig:visual}, to visually compare model performances, we show the CLIP-$T_{(c)}$ (left) for recontextualization and  CLIP-$T_{(t)}$ for texture editing experiments on 30 subjects. The left chart presents the results for the recontextualization task, where the exact subject is placed within a new context described by prompts. Our method consistently outperforms the baselines. The right chart depicts the results for the texture editing task, wherein the texture of the subject itself is altered by prompts. Both baselines experience a dramatic drop in quality. Consequently, the performance gain of our method becomes even more significant.

\subsection{Ablation and Analysis}

In this section, we conduct ablation studies on our proposed subject-cross-attention modules and self-attention masking mechanism. The contextualized features derived from our MLLM predominantly cater to semantic understanding and the general appearance, but they inherently lack fine-grained details. To address this, we introduce a subject-cross-attention feature transfer mechanism, coupled with a masking procedure. This combination serves as a vital component for enhancing detail fidelity. The effectiveness of this approach is ablated below.

\begin{figure}
  \centering
  \includegraphics[width=1.0\textwidth]{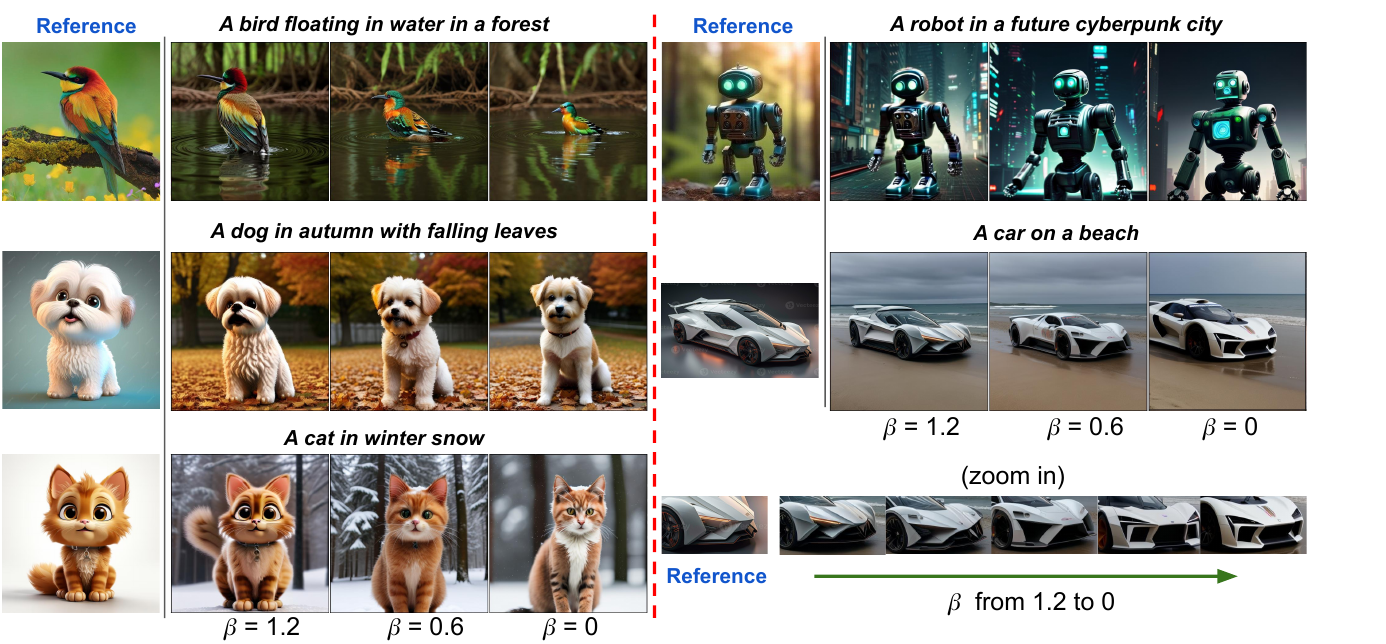} 
  \caption{Effect of self-attention feature transfer. We test various self-attn injection strengths ($\beta$) on the same fixed $Z_T$. Our self-attn transfer is highly effective in keeping exact subject details. As $\beta$ increases from 0 to 1.2, we see results details improve consistently: the cat/dog results change from photorealism to Pixar style, and the robot/bird becomes more accurate. We show zoom-in images for the car and notice how the shape and light of the car headlight gradually become accurate as $\beta$ increases.}
  \label{fig:abl_self}
\end{figure}

\begin{figure}
  \centering
  \includegraphics[width=1.0\textwidth]{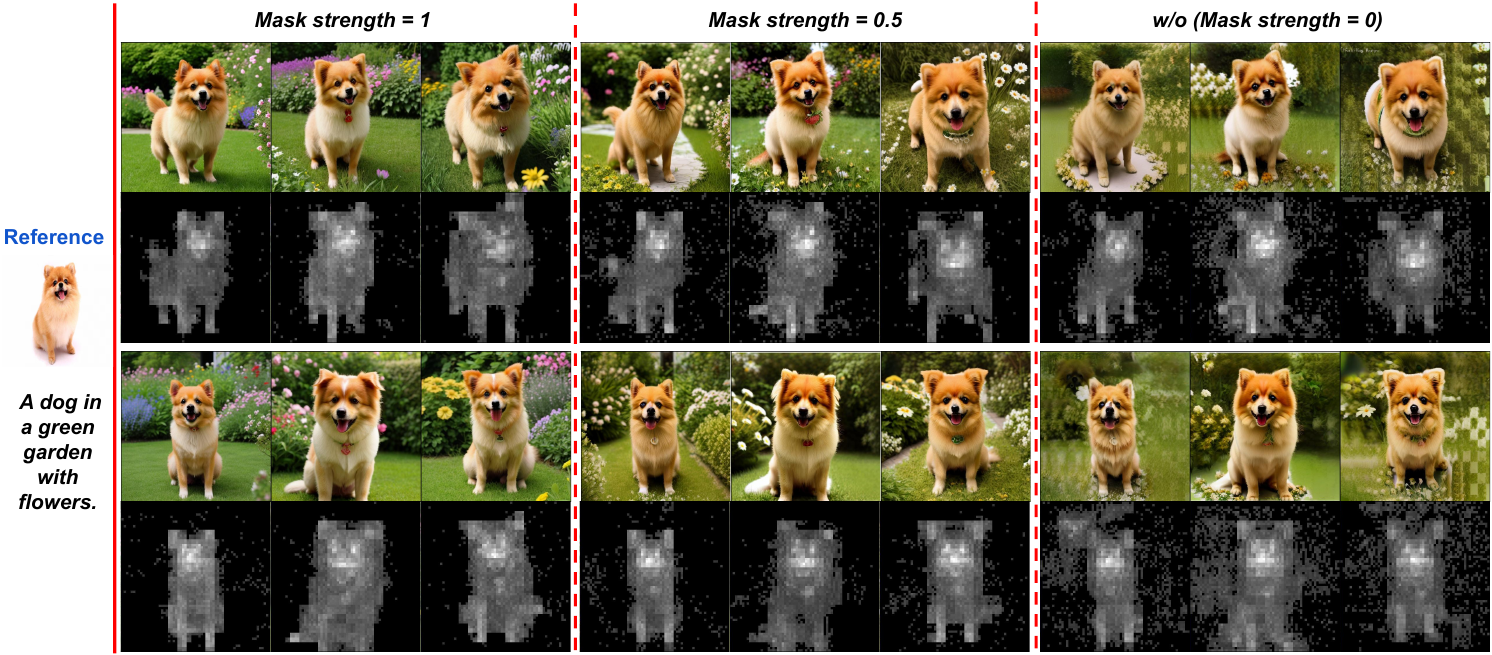} 
  \caption{The effect of self-attention masking. We sample six $Z_T$ noises and monitor their generation results with three mask strengths. As the mask strength decreases, backgrounds get more interfered with, resulting in artifacts. Iterative Masking takes care of the background and foreground separately, greatly improving background quality. The visualization of the cross-attn mask also confirms its effectiveness, as the w/ results are much less diffused than the w/o.}
  \label{fig:abl_mask}
\end{figure}

We show the effectiveness of our attention module's feature transfer ability. In \cref{fig:abl_self}, when $\beta=0$, subject-cross-attention is disabled, and the model entirely relies on the multimodal decoder and its context-cross-attention injection. The generated subjects in the result images are similar to the references in shape and color, but the details are mistaken. As $\beta$ increases, we observe consistent improvement in detail fidelity.

We then analyze the effectiveness of our masking mechanism. As mentioned previously, since $M_{gen}$ is unknown in generation time, the subject-cross-attention masking itself is adapted to an iterative masking manner. As shown in \cref{fig:abl_mask}, this design ensures the generated backgrounds are unaffected by the self-attn feature transfer and greatly improves image quality. We visualize the cross-attn map $M_{gen(t=0)}$ for the keyword $label$ (in this case, dog) after the last denoising step. Notice $M_{gen(0)}$ is noisy and inaccurate when w/o masking. When it is fully activated, $M_{gen(0)}$ becomes clean and clear and the approximation to $M_{gen}$ becomes accurate. The proposed masking technique is crucial and effective to ensure clear and diverse image backgrounds. We show quantitative results for these ablations in \cref{tab:mask} and \cref{tab:self}.

\begin{table}
\centering
\begin{minipage}{0.52\textwidth} 
    \caption{Masking Ablation.}
    \label{tab:mask}
    \resizebox{\textwidth}{!}{
        \begin{tabular}{c|ccc|ccc}
        \hline
        Masking & DINO $\uparrow$ & CLIP-$T_{(c)}$ $\uparrow$ & CLIP-$I$ $\uparrow$ &CLIP-$T_{(t)}$ \\
        \hline
        1.0 & \textbf{0.618} & \textbf{0.348} & \textbf{0.803}  & \textbf{0.335}\\
        0.5 & 0.603 & 0.322 & 0.797 & 0.231\\
        w/o & 0.575 & 0.298 & 0.788 & 0.322  \\
        \hline
        \end{tabular}
    }
\end{minipage}\hfill 
\begin{minipage}{0.48\textwidth} 
    \caption{Subject-cross-attn Ablation.}
    \label{tab:self}
    \resizebox{\textwidth}{!}{
        \begin{tabular}{c|ccc|ccc}
        \hline
        $\beta$ & DINO $\uparrow$ & CLIP-$T_{(c)}$ $\uparrow$ & CLIP-$I$ $\uparrow$ &CLIP-$T_{(t)}$ \\
        \hline
        1.0 & \textbf{0.618} & 0.348 & \textbf{0.803}  & 0.335\\
        0.5 & 0.591 & 0.348 & 0.788 & \textbf{0.338}\\
        w/o & 0.583 & 0.349 & 0.782 & 0.319  \\
        \hline
        \end{tabular}
    }
\end{minipage}\hfill 
\end{table}

\begin{figure}
\centering
\includegraphics[width=0.98\linewidth]{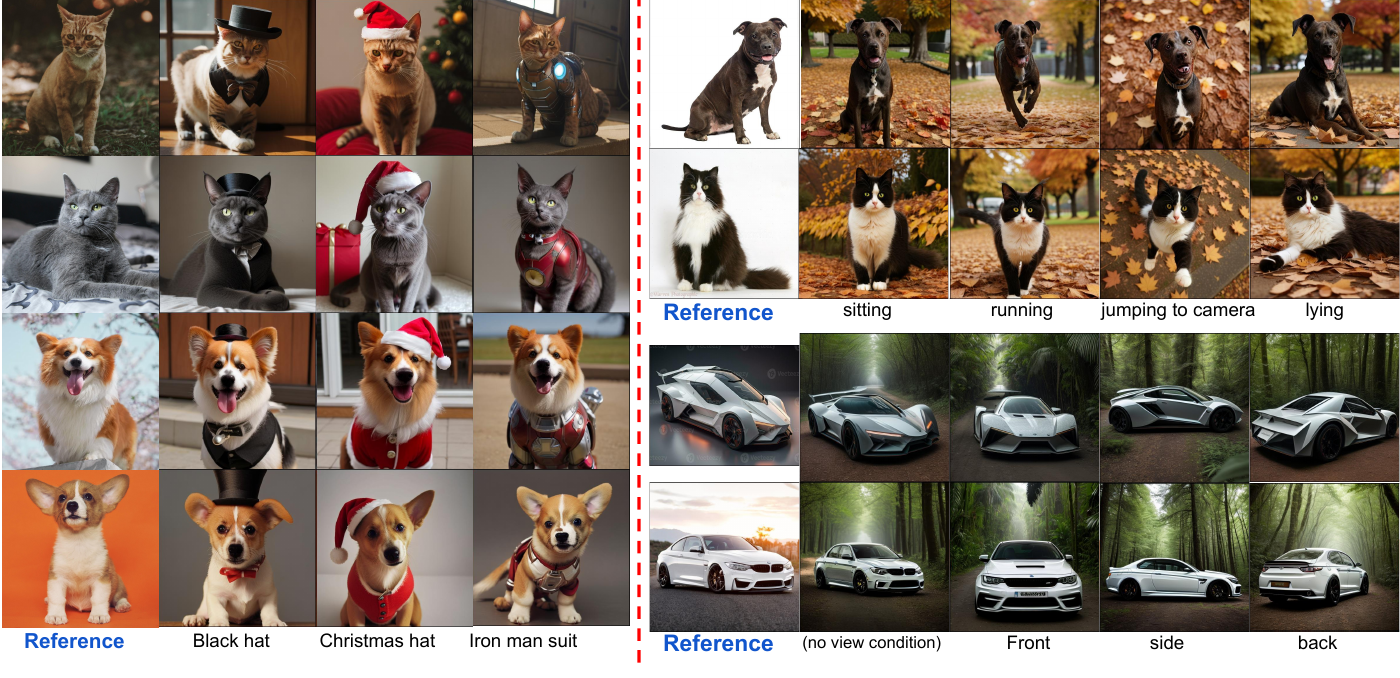} 
\caption{More results on accessory incorporation, pose modification, and camera perspective control.}
\label{fig:exp_1}
\end{figure}

As further analysis, we test our model on more problem settings in addition to the previous results. As shown in \cref{fig:exp_1}, our model generates high-quality subject-coherent images across diverse problem settings, including accessory incorporation, pose modification, and camera perspective controls. This shows it can faithfully generate subject details while allowing flexibility in prompts. 

\begin{figure}
\centering
\includegraphics[width=1.0\linewidth]{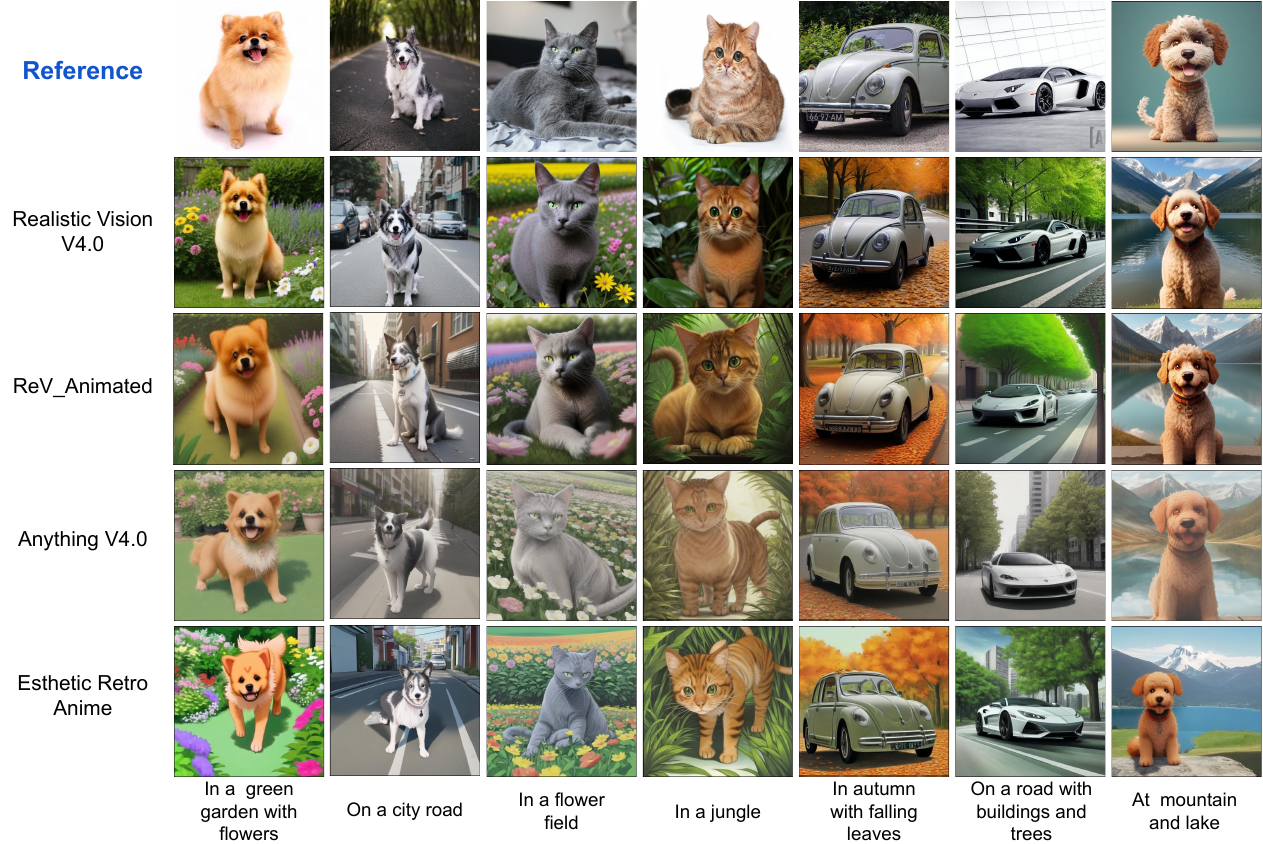} 
\caption{Although trained on RealisticVision, our model can be directly applied to various community checkpoints and yields high-quality coherent results. }
\label{fig:exp_2}
\end{figure}

Additionally, our model is an universal adapter because we freeze the original diffusion model in the training stage. It can generalize to the custom model checkpoints fine-tuned from the same base model. In \cref{fig:exp_2}, we verify this on community models from HuggingFace and CivitAi\cite{huggingface2024civit} including Realistic Vision V4.0\cite{huggingface2024real}, ReV-Animated\cite{huggingface2024rev}, Anything v4\cite{huggingface2024anything} and Esthetic Retro Anime\cite{huggingface2024est}. These models are all fine-tuned from SD v1.5. MoMA can be directly applied to these community models without any modification. 


\section{Conclusion}
In conclusion, we have proposed the powerful MoMA for fast image personalization on the text-to-image diffusion model. It is tuning-free, open-vocabulary, and supports recontextualization and texture editing. The results of our experiments show its superiority against existing methods. Our proposed multimodal image-feature decoder successfully harnesses the strength of MLLM for contextualized feature generation. Our masked subject-cross-attention technique provides a compelling feature shortcut significantly improving detail accuracy. Additionally, as a plug-and-play module, our model can be directly integrated with community models tuned from the same base model, extending its applications to a broader area. 

\clearpage  
\bibliographystyle{splncs04}
\bibliography{egbib}

\clearpage  

\section*{Appendix} 

\section*{A. Implementation Details}

\subsection*{A2. Architecture} 
We leverage the advanced pre-existing knowledge from pretrained MLLMs to aid in model convergence. Specifically, we initialize our multi-modal decoder with LLaVA-v1.5-7B checkpoint\cite{llava7b} from HuggingFace. Its vision encoder is CLIP-vit-large-patch14 (CLIP-L)\cite{clipl} and takes in $I_{ref}^{wb}$. We use ViT-H/1\cite{cliph} to extract the image embedding of $I_{ref}$. Thus, a trained linear mapping is added at the output of MLLM to map the learned text token from $\mathbb{R}^{4096}$ to $\mathbb{R}^{1024}$. The model is built on top of the IP-Adapter\cite{ye2023ip-adapter} and the training code on LDM\cite{rombach2022high}. Our foundation model is StableDiffusion v1.5\cite{rombach2022high} with Realistic-Vision-v4.0\cite{huggingface2024real} checkpoint. Both subject and context cross-attentions are implemented with the diffusers attention processor class. We call our MLLM as multi-modal decoder because: it takes in visual features of $I_{ref}^{wb}$ and text instruction and "generates the full image $I_{ref}$" in the form of CLIP embeddings. So our MLLM is not a typical subject-feature encoder like in most existing works, but instead a generative MLLM decoder.

\subsection*{A2. Training}
we lock the parameters of the image encoders and only adjust the MLLM parameters along with the newly incorporated learnable token. In stage two training, the image/text encoder, multi-modal decoder, and the entirety of the UNet parameters are frozen. We optimize exclusively the linear mappings of subjet-cross-attention and context-cross-attentions.To expedite convergence, we employ attention mapping layers from the IP-adapter\cite{ye2023ip-adapter} and start with a zero-initialized subject-cross-attention. We apply large gradient accumulation steps (200) and clip grad norm for smoother training. In both stages, we use the AdamW optimizer with a learning rate of \( 1^{-5} \). Training takes 7 days in total on 16 A100 80G GPUs. 

\subsection*{A3. Evaluation} 
We use the DDIM sampler with default parameters and 50 denoising steps. Evaluation requires 23.5G GPU RAM for $512\times512$ resolution. The iterative masking is disabled in the first denoising step since $M_{gen(t)}$ is unavailable yet. We apply classifier-free guidance on both the text prompt and Context-cross-attention side, but not the subject-cross-attention. On the text prompt side, simply replace the target prompt $P_{tgt}$ with null embedding. On the context-cross-attention side, we zero out the contextualized features before feeding them to attention linear mappings. We use $\beta=1.0$ in the recontextualization task for better subject detail accuracy and, empirically, $\beta=0.5$ in the texture editing task for better balance between subject and prompt.

\section*{B. Visual comparison and more Results}

\subsection*{B1. Uncurated Comparison with Baselines}
\begin{figure}
\centering
\includegraphics[width=1.0\linewidth]{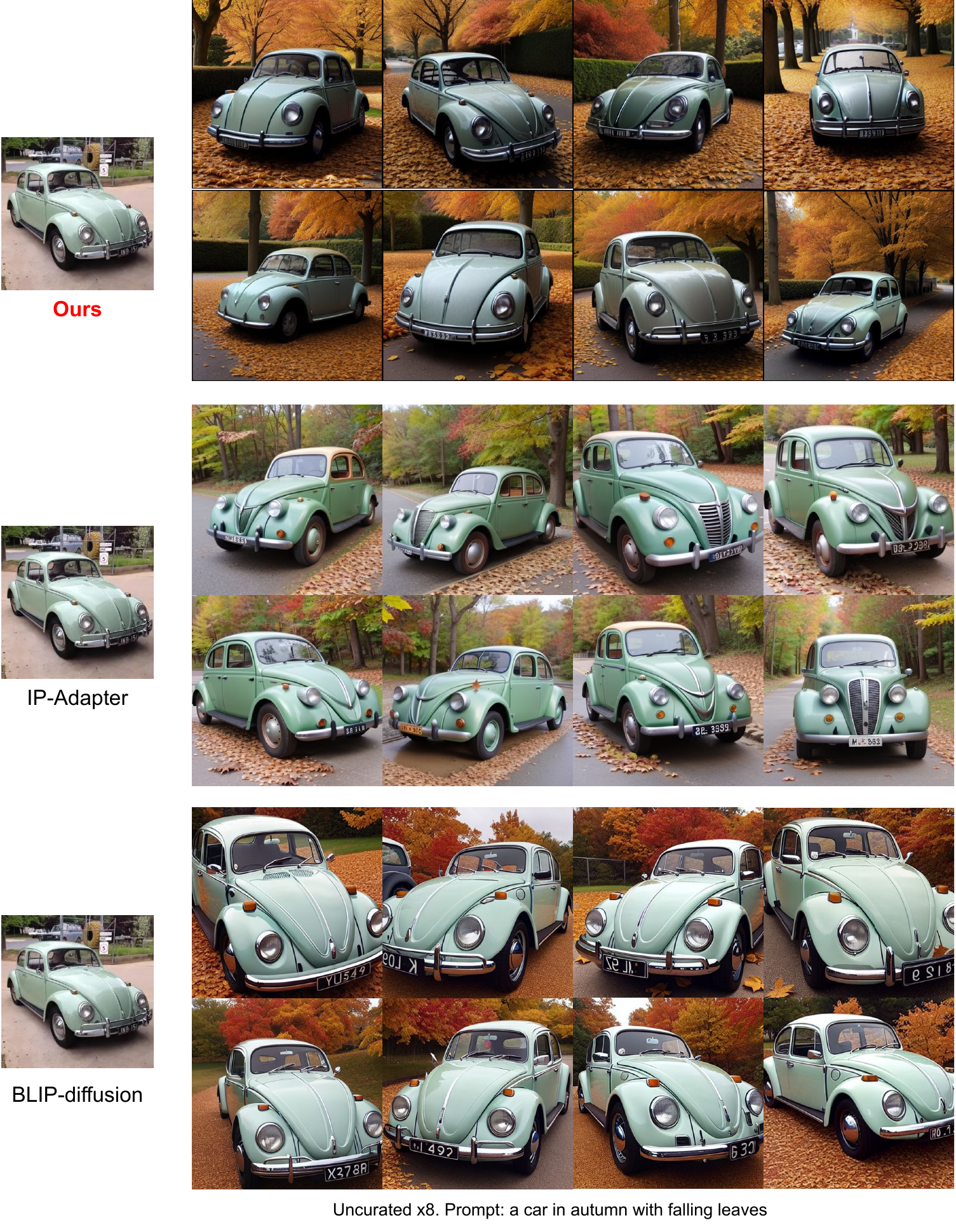} 
\caption{Uncurated random renderings from our model and baselines. IP-Adapter failed in capturing car structures, despite increasing its scale from default 0.6 to 0.85. Our results have better detail accuracy, more diverse composition, and favorable quality than the baselines.}
\label{fig:uncurated_1}
\end{figure}

\begin{figure}
\centering
\includegraphics[width=1.0\linewidth]{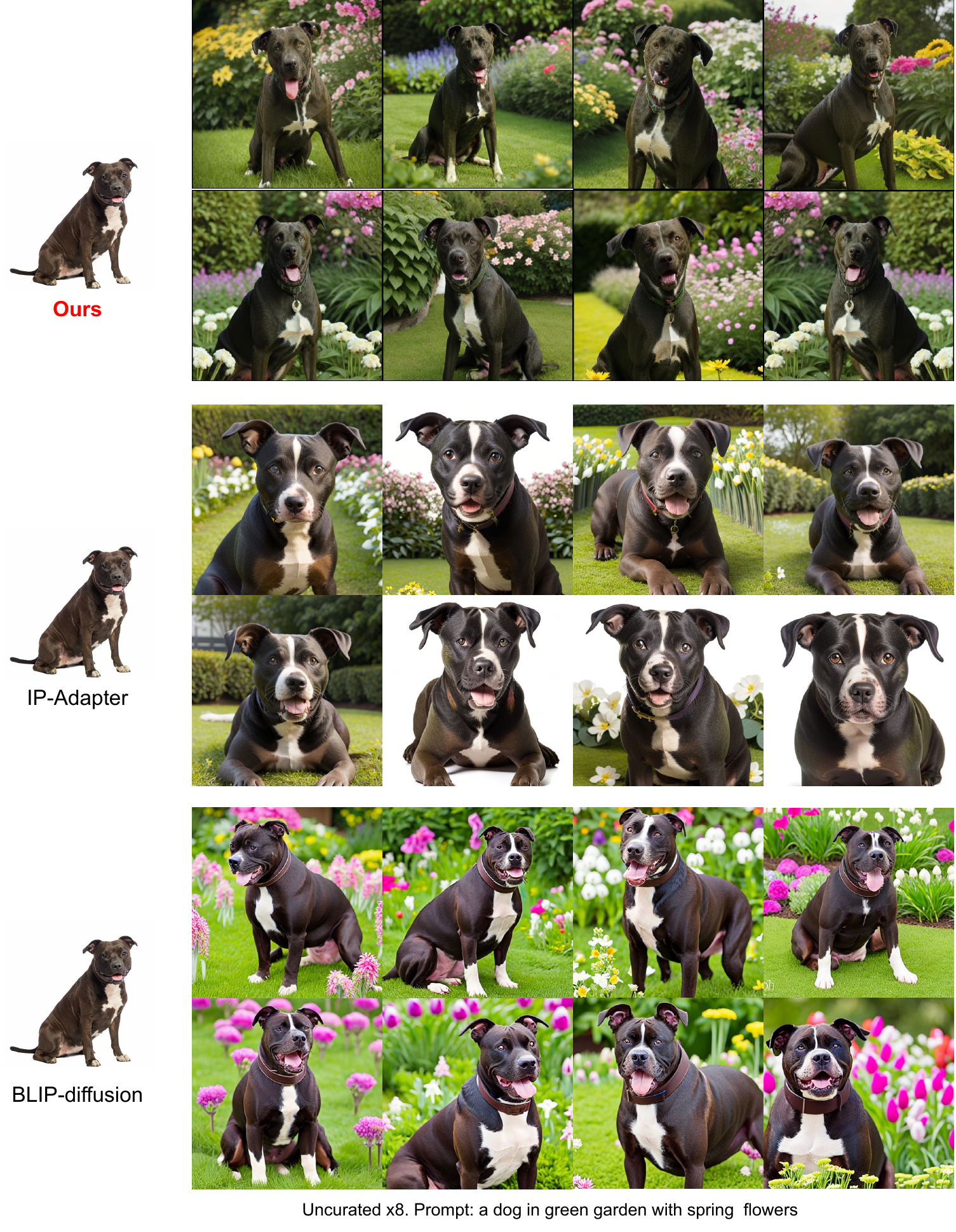} 
\caption{Uncurated random renderings from our model and baselines. IP-adapter generates inaccurate color patterns, even with an increased scale(weight) to 0.85. Since it does not distinguish background from foreground, some results failed to follow the "garden with flower" in the prompt. Our results have more accurate details with much better backgrounds.}
\label{fig:uncurated_2}
\end{figure}

We show uncurated results for our model and the baselines. Results are random and not selected. For each method, 8 images are generated. IP-Adapter failed to accurately capture the structure of the car in \cref{fig:uncurated_1}. In \cref{fig:uncurated_2}, increasing the weight/scale to 0.85 in IP-Adapter still yields a wrong color pattern of the dog. Under this scale, The original white background strongly interferes    with its generated image. Ours correctly captures the color and structure pattern with much more favorable backgrounds, respecting both the reference image and the text prompt.

\begin{table}
\centering
\begin{tabular}{l|cccc|cc|cc}
\toprule
 & \multicolumn{4}{c}{\textbf{Ours}} & \multicolumn{2}{c}{IP-adapter} & \multicolumn{2}{c}{Blip-diffusion} \\
\cmidrule(lr){2-5} \cmidrule(lr){6-7} \cmidrule(lr){8-9}
         & DINO & CLIP-$I$ & CLIP-$T_{(c)}$ &CLIP-$T_{(t)}$ & CLIP-$T_{(c)}$ &CLIP-$T_{(t)}$ & CLIP-$T_{(c)}$ &CLIP-$T_{(t)}$ \\
\midrule
backpack  & 0.590 & 0.861  & 0.350    & 0.338    & 0.342    & 0.337    & 0.324    & 0.292 \\
backpack2 & 0.438 & 0.698  & 0.361    & 0.351    & 0.357    & 0.365    & 0.323    & 0.296 \\
boot & 0.502 & 0.819 & 0.346 & 0.345 & 0.330 & 0.325 & 0.287 & 0.277 \\
bowl & 0.621 & 0.699 & 0.350 & 0.338 & 0.300 & 0.339 & 0.249 & 0.257 \\
can & 0.657 & 0.749 & 0.361 & 0.321 & 0.355 & 0.347 & 0.310 & 0.294 \\
candle & 0.474 & 0.734 & 0.361 & 0.353 & 0.348 & 0.330 & 0.329 & 0.297 \\
cartoon & 0.574 & 0.794 & 0.305 & 0.312 & 0.284 & 0.278 & 0.255 & 0.248 \\
cat & 0.813 & 0.840 & 0.354 & 0.339 & 0.330 & 0.300 & 0.320 & 0.261 \\
cat2 & 0.750 & 0.845 & 0.358 & 0.345 & 0.327 & 0.310 & 0.292 & 0.263 \\
clock & 0.679 & 0.866 & 0.335 & 0.341 & 0.305 & 0.324 & 0.322 & 0.270 \\
dog & 0.725 & 0.863 & 0.342 & 0.334 & 0.328 & 0.308 & 0.302 & 0.261 \\
dog2 & 0.652 & 0.863 & 0.338 & 0.335 & 0.322 & 0.301 & 0.302 & 0.260 \\
dog3 & 0.609 & 0.799 & 0.342 & 0.345 & 0.326 & 0.304 & 0.303 & 0.257 \\
dog5 & 0.574 & 0.761 & 0.344 & 0.340 & 0.298 & 0.279 & 0.324 & 0.257 \\
dog6 & 0.713 & 0.857 & 0.332 & 0.324 & 0.320 & 0.300 & 0.282 & 0.253 \\
dog7 & 0.709 & 0.850 & 0.338 & 0.338 & 0.325 & 0.287 & 0.305 & 0.248 \\
dog8 & 0.654 & 0.842 & 0.337 & 0.333 & 0.316 & 0.289 & 0.306 & 0.246 \\
plushie & 0.620 & 0.781 & 0.367 & 0.334 & 0.341 & 0.315 & 0.304 & 0.287 \\
plushie2 & 0.491 & 0.754 & 0.372 & 0.333 & 0.340 & 0.336 & 0.326 & 0.299 \\
plushie3 & 0.621 & 0.794 & 0.366 & 0.323 & 0.343 & 0.317 & 0.289 & 0.271 \\
poop & 0.533 & 0.732 & 0.345 & 0.339 & 0.335 & 0.338 & 0.327 & 0.291 \\
sneaker & 0.665 & 0.814 & 0.350 & 0.347 & 0.333 & 0.312 & 0.276 & 0.256 \\
sneaker2 & 0.704 & 0.824 & 0.340 & 0.346 & 0.325 & 0.305 & 0.269 & 0.264 \\
sunglasses & 0.655 & 0.851 & 0.349 & 0.344 & 0.339 & 0.348 & 0.299 & 0.293 \\
teapot & 0.642 & 0.852 & 0.383 & 0.361 & 0.366 & 0.359 & 0.294 & 0.288 \\
toy1 & 0.508 & 0.765 & 0.341 & 0.306 & 0.332 & 0.292 & 0.292 & 0.258 \\
toy2 & 0.627 & 0.821 & 0.334 & 0.310 & 0.320 & 0.328 & 0.287 & 0.260 \\
toy3 & 0.532 & 0.741 & 0.348 & 0.313 & 0.337 & 0.311 & 0.292 & 0.266 \\
vase & 0.593 & 0.815 & 0.358 & 0.339 & 0.338 & 0.356 & 0.295 & 0.299 \\
Average & 0.618 & 0.803 & 0.348 & 0.335 & 0.330 & 0.319 & 0.300 & 0.271 \\
\bottomrule
\end{tabular}
\caption{Quantitative details on DreamBooth datasets. 
We add 10 new prompts focusing on texture editing to calculate CLIP-$T_{(t)}$. These prompts are: A sculpture of a $label$ made of (Lego/Paper/Gold/Wood/silver/green jade/glass/stone/sketch/Minecraft).
}
\label{tab:my_label}
\end{table}

We conclude from above that:(i) Our multi-modal LLM decoder and its contextualized feature are crucial to our model performance. It extracts visual features from reference image $I_{ref}^{wb}$ and contextualizes it with the target text prompt $P_{tgt}$. Its output is trained to match the overall CLIP image embedding of the resulting image and provides critical features to bridge the gap between the subject feature and the target prompt. (ii) Iterative masking helps distinguish between the background and main subject, improving CLIP-T. Removing it leads to corrupted backgrounds and significantly lower image quality. (iii) Disabling the Subject-cross-attention worsens the CLIP-I score as features from MLLM are insufficient to accurately reproduce subject details. This quantitative ablation lines up with our intuition: our multi-modal decoder and its context-cross-attention provide vital image features and serve as a foundation. Subject-cross-attention and iterative masking help in subject detail accuracy and background quality respectively.


\subsection*{B3. More Results}
\begin{figure}
\centering
\includegraphics[width=1.0\linewidth]{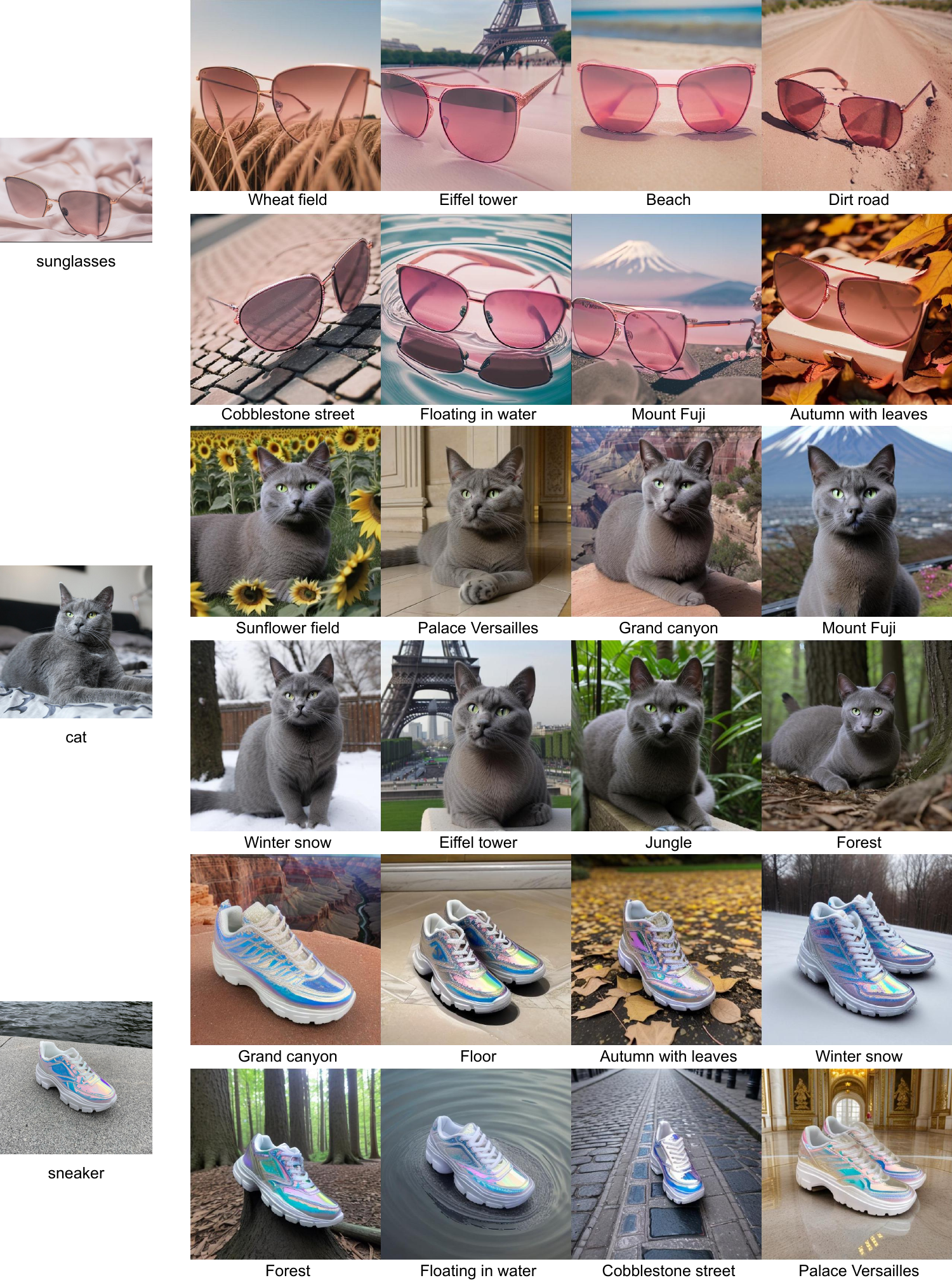} 
\caption{Additional results from our model. Change context.}
\label{fig:appendix1}
\end{figure}

\begin{figure}
\centering
\includegraphics[width=1.0\linewidth]{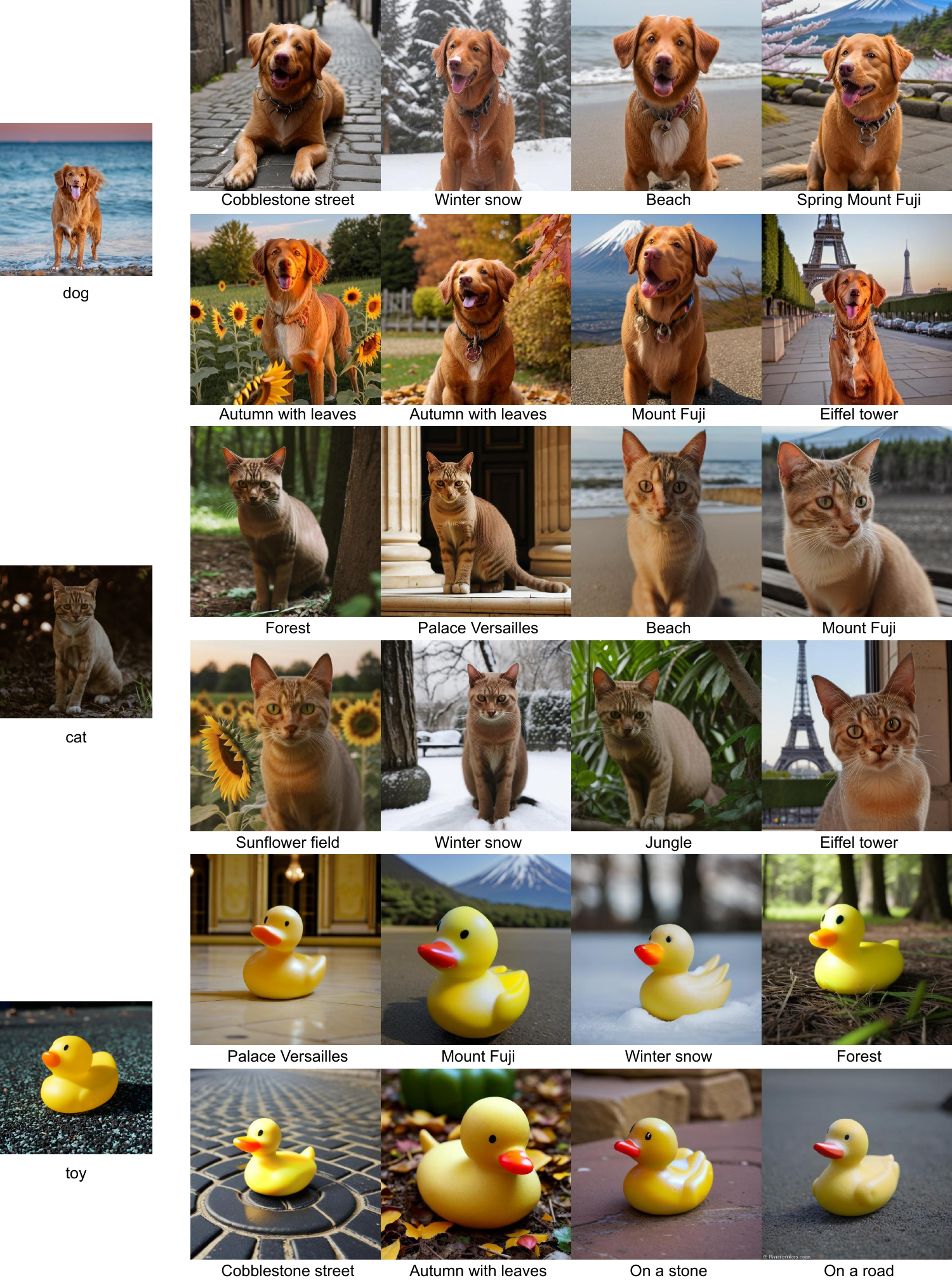} 
\caption{Additional results from our model. Change context.}
\label{fig:appendix2}
\end{figure}

\begin{figure}
\centering
\includegraphics[width=1.0\linewidth]{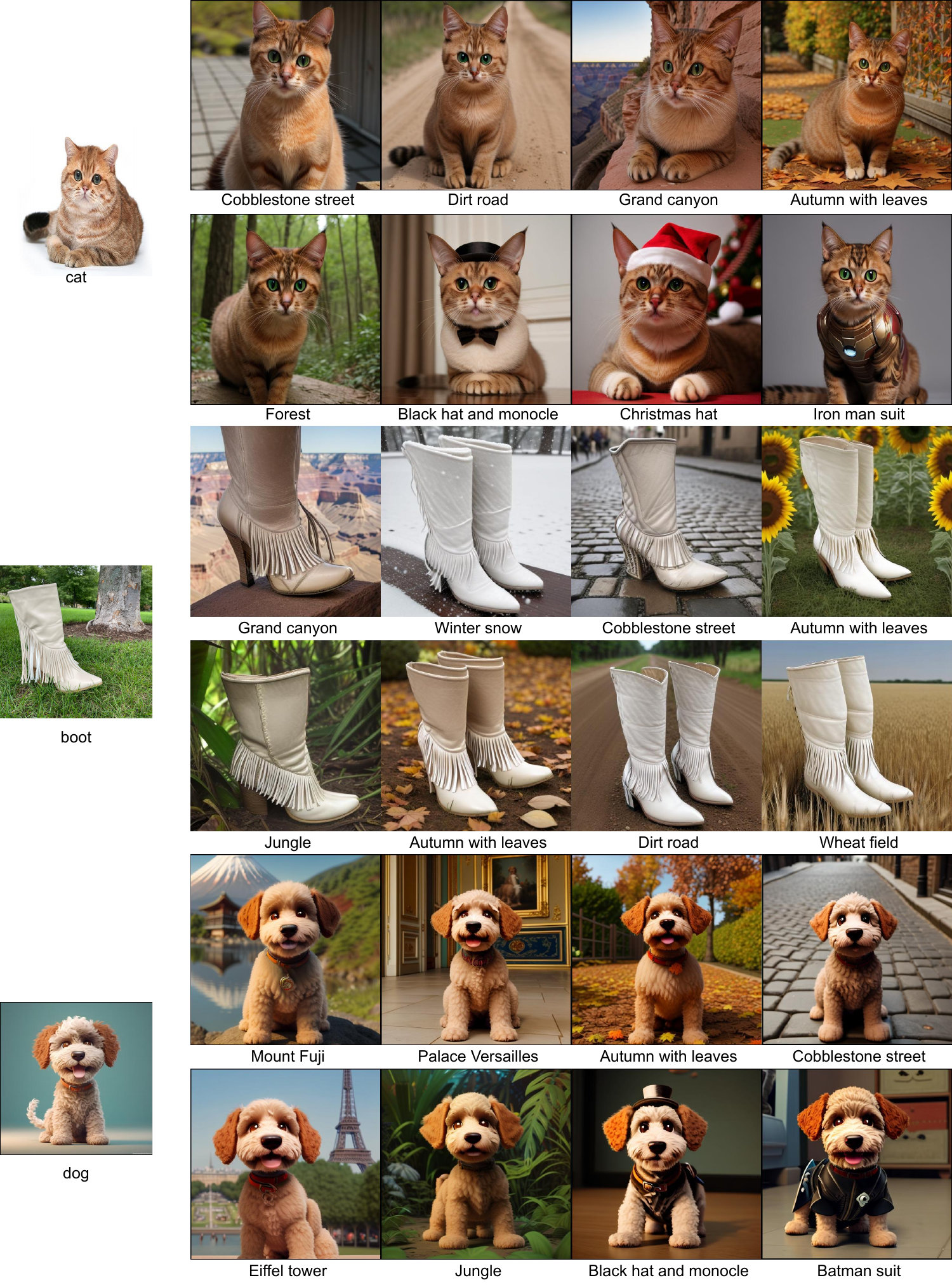} 
\caption{Additional results from our model. Change context.}
\label{fig:appendix3}
\end{figure}

\begin{figure}
\centering
\includegraphics[width=1.0\linewidth]{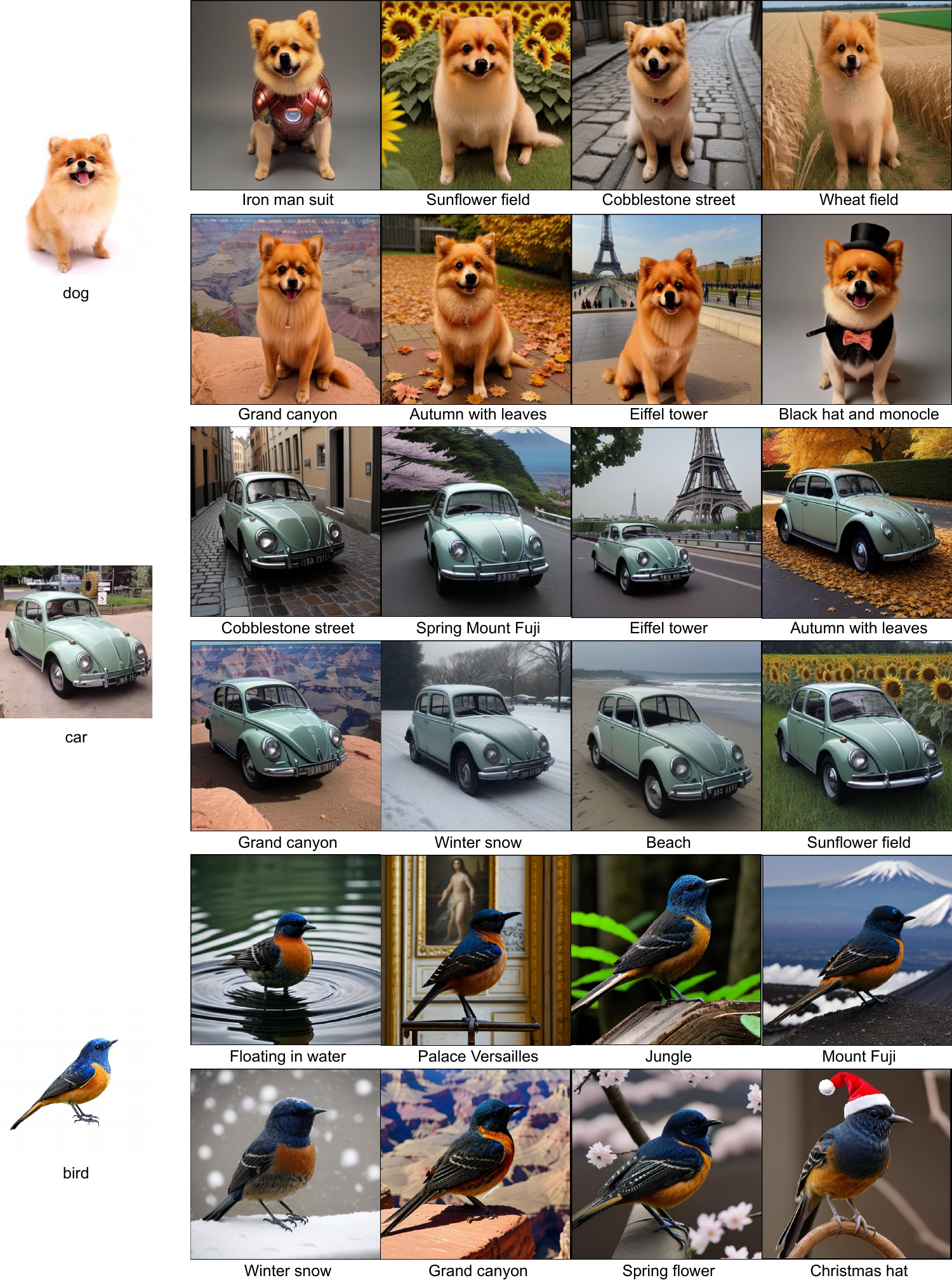} 
\caption{Additional results from our model. Change context.}
\label{fig:appendix4}
\end{figure}

\begin{figure}
\centering
\includegraphics[width=1.0\linewidth]{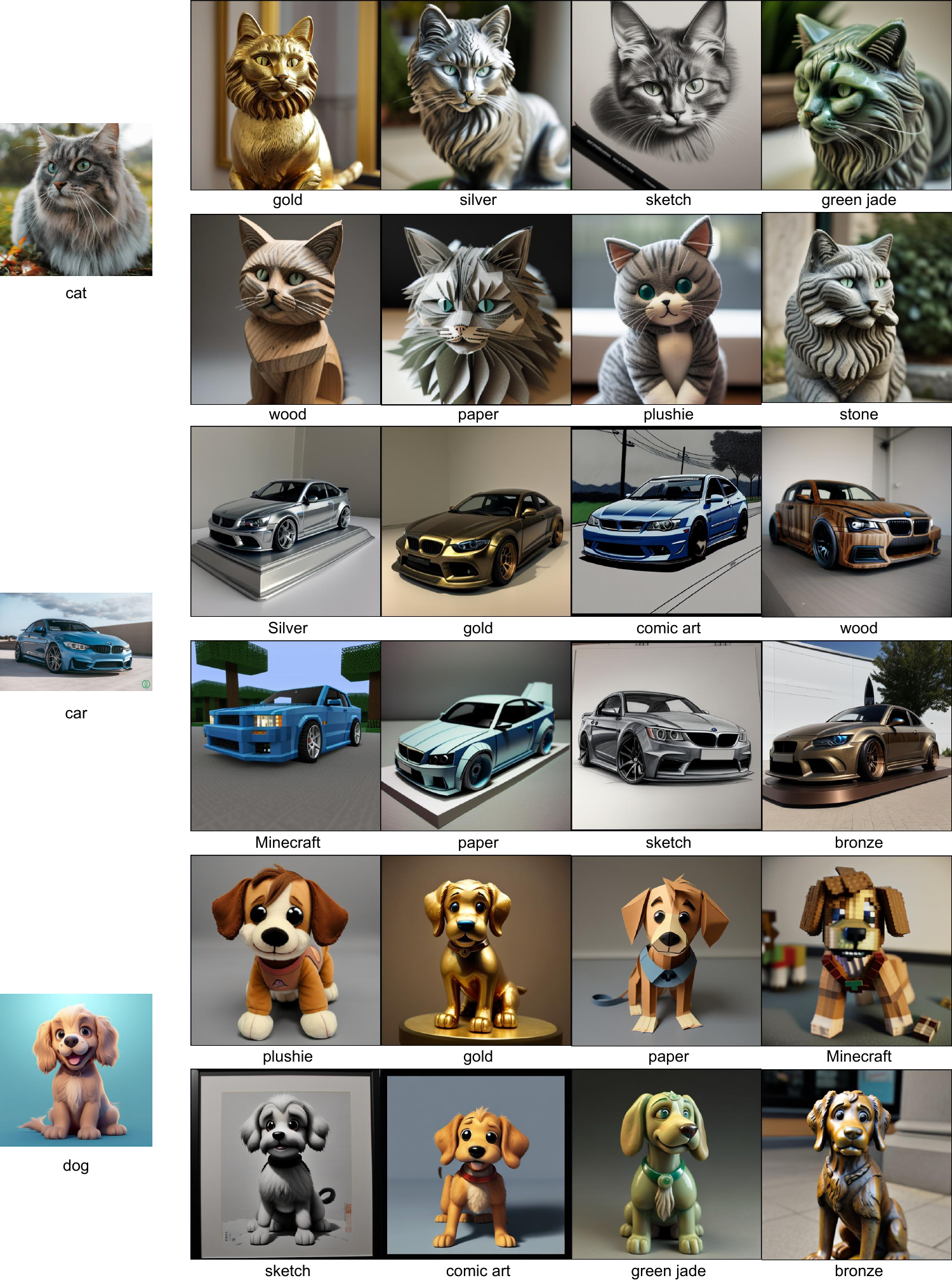} 
\caption{Additional results from our model. Change texture.}
\label{fig:appendix5}
\end{figure}

\begin{figure}
\centering
\includegraphics[width=1.0\linewidth]{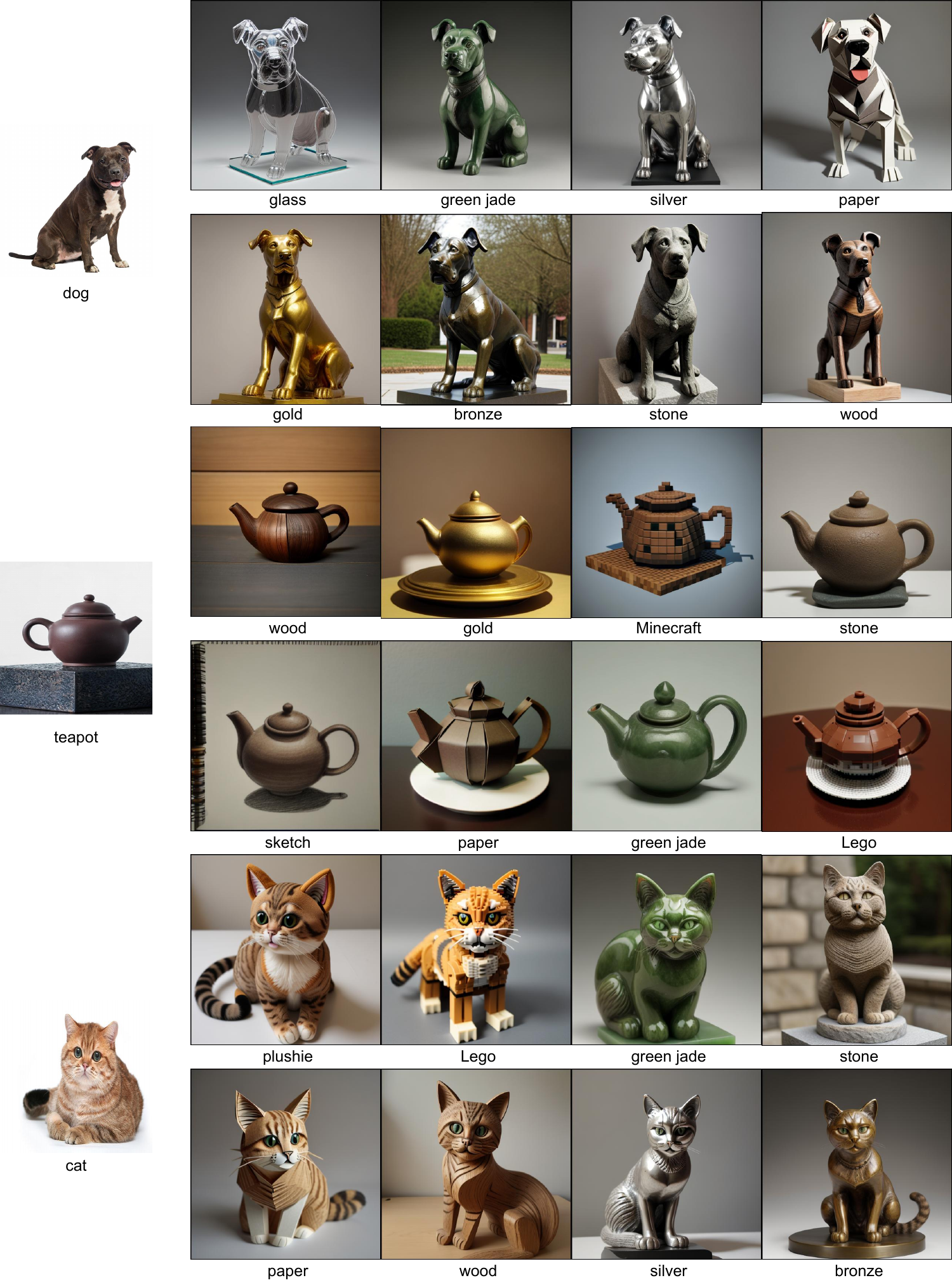} 
\caption{Additional results from our model. Change texture.}
\label{fig:appendix6}
\end{figure}

In \cref{tab:my_label}, we show detailed quantitative results for the DreamBooth dataset. From \cref{fig:appendix1} to \cref{fig:appendix6}, we provide additional qualitative results on various subjects and prompts. Reference subject images are on the left. In the rest columns, we provide generated renditions. \cref{fig:appendix1} to \cref{fig:appendix4} are about context editing and \cref{fig:appendix5} and \cref{fig:appendix6} are about texture editing.

\section*{C. Discussions}
\subsection*{C1. Social Impact}

While tuning-based personalization models are largely inaccessible to most people because of computation resource limits, our method of open-vocab tuning-free personalization model helps democratize such models to everyday users with a significantly improved quality. However, it also bears the potential risk of being exploited for the creation of deceptive content or the propagation of misinformation. To address this concern, we have specifically designed our training process to exclude person-related subjects and focus on generic objects. This intentional limitation reduces the model's ability to generate convincing counterfeit images where individuals are central elements. To ensure the integrity of content generated by our model, we advise a thorough examination of its outputs before deploying our model in consumer-facing applications.  

\subsection*{C2. Failure Examples}

\begin{figure}
\centering
\includegraphics[width=1.0\linewidth]{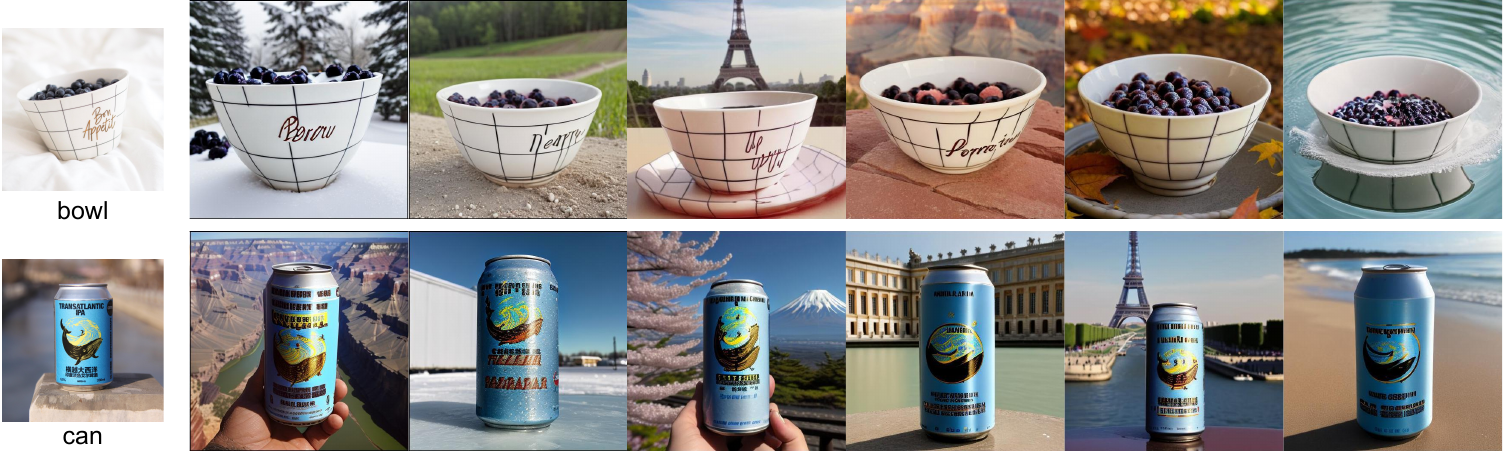} 
\caption{Failure cases. The \textbf{text} messages displayed on the bowl and can appear distorted or missing in the resulting images, a flaw inherited from SD v1.5. This version is notably challenged in its ability to reproduce text with accuracy.}
\label{fig:failed}
\end{figure}

We also noticed that some subjects are much easier to learn than others\cite{li2024blip}. For example, the model generates high-quality results for dogs and cats with consistent identity and almost identical details. Our improvement against baselines\cite{li2024blip}\cite{ye2023ip-adapter} starts to be more noticeable for rare subjects like shoes and robots. Occasionally, as shown in \cref{fig:failed}, with subjects that are rarer especially accompanied by text, the model is unable to fully capture its details.

\subsection*{C3. About User Study}
We design 9 questions: 6 for recontextualization task (3 about subject fidelity, 3 about background-prompt fidelity) and 3 for texture editing task. Nine ratings per user and a total of 774 ratings were collected. A sample is shown below:

\begin{figure}
\centering
\includegraphics[width=0.9\linewidth]{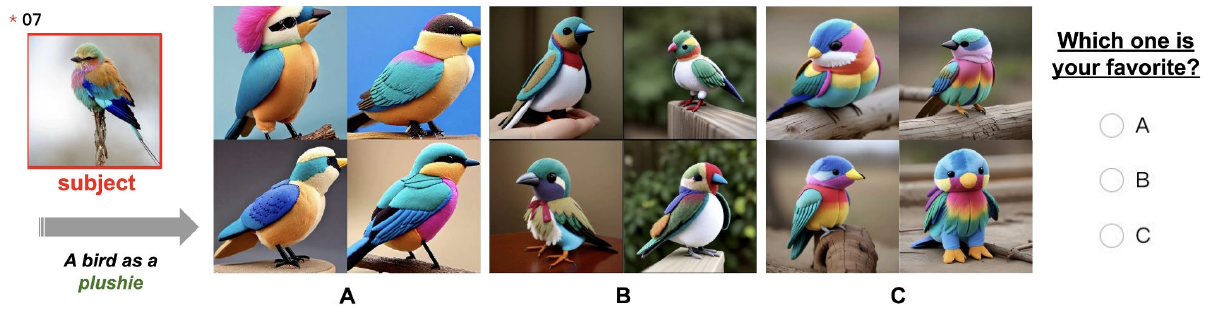} 
\label{fig:user}
\end{figure}

\clearpage



%
%

\end{document}